\documentclass{article}

\usepackage{microtype}
\usepackage{graphicx}
\usepackage{subfigure}
\usepackage{placeins}

\usepackage{booktabs} 
\usepackage{multirow}
\usepackage{multicol}

\usepackage{hyperref}
\usepackage{rotating} 

\usepackage[accepted]{icml2024}

\usepackage{amsmath}
\usepackage{amssymb}
\usepackage{mathtools}
\usepackage{amsthm}

\usepackage[capitalize,noabbrev]{cleveref}

\usepackage{dblfloatfix}

\usepackage{float}

\usepackage{xurl} 

\usepackage{hyperref}

\theoremstyle{plain}
\newtheorem{theorem}{Theorem}[section]

\theoremstyle{definition}

\newtheorem{assumption}[theorem]{Assumption}
\theoremstyle{remark}

\include{some_math.sty}
\usepackage[textsize=tiny,disable]{todonotes}

\newcommand{\ih}[1]{\todo[inline]{[IH]: #1}}
\newcommand{\jp}[1]{\todo[inline]{[JP]: #1}}

\icmltitlerunning{Loss Shaping Constraints for Long-Term Forecasting}

\begin{document}

\twocolumn[
\icmltitle{Loss Shaping Constraints for Long-Term Time Series Forecasting}

\icmlsetsymbol{equal}{*}

\begin{icmlauthorlist}
\icmlauthor{Ignacio Hounie}{equal,upenn}
\icmlauthor{Javier Porras-Valenzuela}{equal,upenn}
\icmlauthor{Alejandro Ribeiro}{upenn}
\end{icmlauthorlist}

\icmlaffiliation{upenn}{University of Pennsylvania}

\icmlcorrespondingauthor{Ignacio Hounie}{ihounie@seas.upenn.edu}

\icmlkeywords{Machine Learning, Time Series Prediction, Multi-step, Loss Shaping, Constrained Optimization, Deep Learning, Transformers, Duality.}
\vskip 0.3in
]


\printAffiliationsAndNotice{\icmlEqualContribution} 

\begin{abstract}
Several applications in time series forecasting require predicting multiple steps ahead. Despite the vast amount of literature in the topic, both classical and recent deep learning based approaches have mostly focused on minimising  performance~\emph{averaged} over the predicted window. We observe that this can lead to disparate distributions of errors across forecasting steps, especially for recent transformer architectures trained on popular forecasting benchmarks. 
That is, optimising performance on average can lead to undesirably large errors at specific time-steps.
In this work, we present a Constrained Learning approach for long-term time series forecasting that aims to find the best model in terms of average performance that respects a user-defined upper bound on the loss at each time-step. We call our approach~\emph{loss shaping constraints} because it imposes constraints on the loss at each time step, and leverage recent duality results to show that despite its non-convexity, the resulting problem has a bounded duality gap. We propose a practical Primal-Dual algorithm to tackle it, and demonstrate that the proposed approach exhibits competitive average performance in time series forecasting benchmarks, while shaping the distribution of errors across the predicted window. 
\end{abstract}

\begin{figure}[t]
        \centering
         \includegraphics[width=0.9\linewidth]{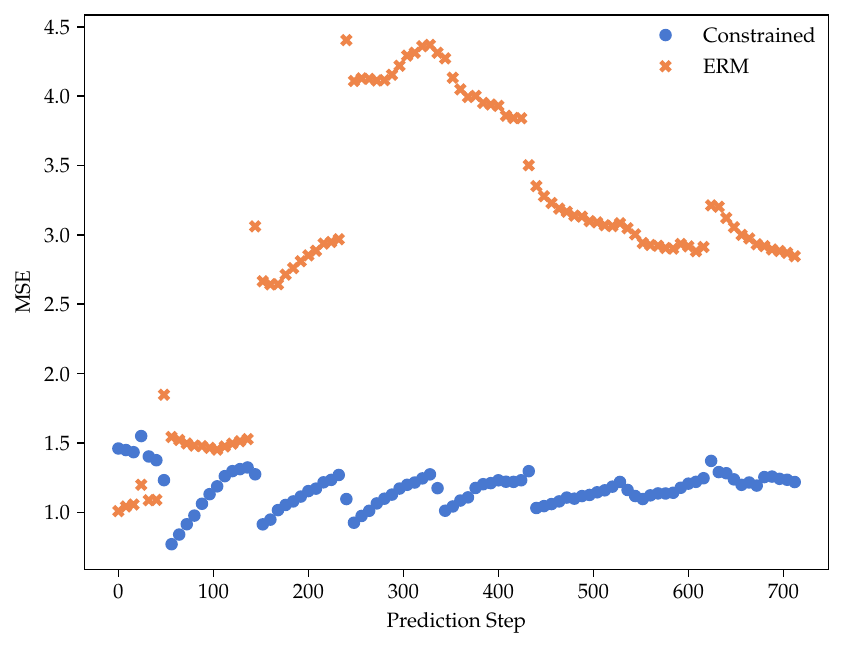}
    \caption{Test Mean Squared Error (MSE) computed at individual time steps on the forecasting window for Autoformer~\cite{wu2021autoformer} on exchange rate data using ERM and our approach.}\label{fig:intro}
\end{figure}

\section{Introduction}\label{sec:intro}

Predicting multiple future values of time series data, also known as multi-step forecasting~\cite{book-survey-multi-step-ML}, has a myriad of applications such as forecasting weather~\cite{multi-step-wind}, electricity demand~\cite{yi2022electric}, prices~\cite{oil-price}, and passenger demand ~\cite{passenger-demand}. Several approaches to generating predictions for the next window have been proposed, including direct~\cite{direct-chevillon, direct-2}, autoregressive or recursive~\cite{recursive-0, recursive-1, recursive-2} and MIMO techniques~\cite{mimo-0, mimo-1, mimo-2, kitaev2019reformer,2021informer, wu2021autoformer, 2021pyraformer,2022fed-former,2022patchTST-transformer,dlinear,2023timegpt, 2023google-foundation, liu2024itransformer, liu2022nonstationary}.
Moreover, a plethora of learning parametrizations exist, ranging from linear models~\cite{boxjenkins1976, dlinear} to recent transformer
~\cite{kitaev2019reformer,2021informer, wu2021autoformer, 2021pyraformer,liu2024itransformer,liu2022nonstationary} and custom~\cite{2022FiLM} architectures.

Regardless of the model and parametrization, most approaches optimize a performance, risk or model fit functional (usually MSE),~\emph{averaged} over the predicted window. Therefore, the distribution of errors across the window -- without any additional assumptions -- can vary depending on the model and data generating process. In practice, this can lead to uneven performance across the different steps of the window. 

Recent works using transformer based architectures focus on aggregate metrics, see for example~\cite{kitaev2019reformer, wu2021autoformer, 2022patchTST-transformer}, while addressing errors at different time steps has recieved little attention~\cite{2023gmm}. 

However, having no control on the error incurred at each time step in the predicted window can be detrimental in scenarios that can be affected by this variability. For example, analysing average behaviour is insufficient to assess financial risks in econometrics ~\cite{europeanbank-quantileVAR} or to ensure stability when using the predictor in a Model Predictive Control framework~\cite{lstm-mpc-stability}.

Therefore, our first contribution is addressing this problem through constrained learning.
\begin{itemize}
\item[(C1)] We formulate multi-step series forecasting as a constrained learning problem, which aims to find the best model in terms of~\emph{average performance} while imposing a user-defined upper bound on the loss at each time-step.
\end{itemize} 
Since this constrains the distribution of the loss per time step across the window, we call this approach \emph{loss shaping constraints}. Due to the challenges in finding appropriate constraints, we leverage a recent approach~\cite{hounie2023resilient} to re-interpret the specification as a soft constraint and find a minimal norm relaxation while jointly solving the learning task. We leverage recent duality results~\cite{chamon2020NIPS} to show that, despite being non-convex, the resulting problem has a bounded duality gap. This allows us to provide approximation guarantees at the step level.

We then propose an alternating Primal-Dual algorithm to tackle the loss shaping constrained problem. Our final contribution is the experimental evaluation of this algorithms in a practical setting:

\begin{itemize}
\item[(C2)] We evaluate our algorithms for different constraints using state-of-the-art transformer architectures~\cite{kitaev2019reformer,2021informer, wu2021autoformer} in popular forecasting benchmarks.
\end{itemize}
Our empirical results showcase the ability to alter the shape of the loss by introducing constraints, and how this can lead to better performance in terms of standard deviation across the predictive window.

\subsection{Related work}

There is little prior work in time series prediction on promoting or imposing a certain distribution of errors \emph{across} the predicted window, i.e., at specific timesteps. Works that propose reweighting the errors at different steps of the window mostly aim to improve average performance by leveraging the structure of residuals~\cite{direct-chevillon, non-linear-reweighting, multi-step-model-selection} and rely on properties of the data generating process and predictive model class. On the other hand, works that address the empirical distribution of forecasting errors and robustness in non-parametric models~\cite{robust-1, robustness-2} do not analyse the distribution of errors \emph{across} multiple steps. Similarly, Multiple output Support Vector Regression for multi-step forecasting~\cite{m-SVR-TS}, which also tackles a constrained problem, has only addressed aggregate errors across the whole window.

Although our method is related to other loss shaping approaches \cite{rebuttal_ref_1, rebuttal_ref_2, 2023gmm} in the sense that it alters the weighting given to each time step on the loss, both the motivation and the problems that these approaches address differ from our work. \citep{rebuttal_ref_1} aim to downweight the impact of large errors as long as they have low probability, regardless of their position in the predictive window. Our approach aims to control average errors at each step and can thus go in the opposite direction of sacrificing performance in order to reduce undesirably large errors at specific steps. ~\cite{rebuttal_ref_2} gives each time-step a fixed weighting that is inversely proportional to their position in the window, thus aiming to discount the impact of far-future errors and give a higher weighting to closer steps. However, imposing a fixed weighting does not take into account the difficulty of fitting an error at a certain time-step, which will depend also on the model and the data. That is, imposing a certain weighting across the window \emph{need not} result in the desired distribution of errors.

Lastly, recent works in generative time series models have also sought to impose constraints using penalty based methods~\cite{2023generativeconstrained}, but the nature of the constraints and proposed approach also differs from ours.

\section{Multi-Step Time Series Forecasting}\label{sec:erm}
Let $\x_t \in \mathcal{X} \subseteq \mathbb{R}^{d_x}$ denote the in feature vector and $ y_t \in \mathcal{Y} \subseteq \mathbb{R}^{d_y}$ its associated output or measurement at time step $t$. The goal in multi-step time series forecasting is to predict $T_{p}$ future values of the output, i.e., $y_{p} := y_{[t+1:t+T_{p}]}$ given a window of length $T_{c}$ of input features, i.e., $\x_{c}:=\x_{[ t - T_{c}:t]}$. The quality of the forecasts can be evaluated at each time-step using a non-negative loss function or metric $\ell: \mathcal{Y} \times \mathcal{Y} \to \mathbb{R}_+$, e.g., the squared error or absolute difference.

The most common approach in supervised multi-step forecasting is to learn a predictor that minimizes the expected loss \emph{averaged over the predicted window}:
\begin{align}\label{Avg-SRM}~\tag{ERM}
\min_{\bbtheta \in \Theta} ~ \mathbb{E}_{(\x_{c}, y_{p} ) \sim \mathfrak{D}}\left[\frac{1}{T_{p}} \sum_{i=1}^{T_{p}} \ell_i({f_{\bbtheta}(\x_{c})}, y_{p})\right],
\end{align}
where~$\ell_i := \ell({[f_{\bbtheta}(\x_{c})}]_i, [y_{p}]_{i})$ denotes the loss evaluated at the $i$-th time step, $\mathfrak{D}$ a probability distribution over data pairs\footnote{Our approach is distribution agnostic, i.e. we impose no additional structure or assumptions on $\mathfrak{D}$.} ${(\x_{c}, y_{p})}$ and ${f_{\bbtheta}: \mathcal{X}^{T_{c}}\to\mathcal{Y}^{T_{p}}}$ is a predictor associated with parameters~$\bbtheta \in \Theta \subset \mathbb{R}^k$, for example, a transformer based neural network architecture.

However, this choice of objective does not account for the \emph{structure} or distribution of the errors across different time steps, which can lead to disparate behaviour across the predicted window as depicted in Figure~\ref{fig:intro}. In particular, we observe empirically that state-of-the-art transformer architectures~\cite{kitaev2019reformer, 2021informer, wu2021autoformer, 2021pyraformer, 2022FiLM} can yield highly varying loss dynamics, including non-monotonic, flat and highly non-linear landscapes, as presented in Appendix~\ref{app-exp-results}
\ih{Add old shape of ERM section to appendix}

In order to control or promote a desirable loss pattern, a weighted average across time steps could be employed. However, since the realised losses will depend not only on the data distribution but also on the model class and learning algorithm. As a result, such penalization coefficients would have to be tuned in order to achieve a desired loss pattern, which in contrast can be naturally expressed as a requirement, as presented next.

\section{Loss Shaping Constraints}\label{sec:Constrained}
In order to control the shape of the loss over the time window, we require the loss on timestep $i$ to be smaller than some quantity $\bbepsilon_i$. This leads to the constrained statistical learning problem:

\begin{tcolorbox}
\vspace{-0.15in}
\begin{align*}
\label{P-LSC}
 P^{\star}(\bbepsilon)=& \min_{\bbtheta \in \Theta}  \; \mathbb{E}_{(\x_{c}, y_{p}) \sim \mathfrak{D}}\left[\frac{1}{T_{p}} \sum_{i=1}^{T_{p}} \ell_i({f_{\bbtheta}(\x_{c})}, y_{p}) \right] \\
&\text{s. to :} \quad \mathbb{E}_{(\x_{c}, y_{p}) \sim \mathfrak{D}}\left[\ell_i({f_{\bbtheta}(\x_{c})}, y_{p})\right] \leq \bbepsilon_i \; \nonumber\\
& \quad\quad i=1,\ldots, T_{p}. \tag{P-LS}\end{align*}
\end{tcolorbox}

An advantage of~\eqref{P-LSC} is that it is interpretable in the sense that constraints -- unlike penalty coefficients -- make explicit the requirement that they represent. That is, the constraint is expressed of terms of the \emph{expected} loss at individual time-steps, and thus prior knowledge about and the underlying data distribution and model class can be exploited.

In Section~\ref{sec:exps}, we focus on simply imposing a constant upper bound for all time steps, which we set using the performance of the (unconstrained) ERM solution. By upper bounding the error, we prevent errors from being undesirably large irrespectively of their position in the window, effectively limiting the spread of the errors across the whole window. Since we still minimize the mean error across the window, this approach is not as conservative as minimax formulations~\cite{minimax}, which only focus on the worst error.

It is worth pointing out that many other constraint choices are possible. For instance, it is often reasonable to assume errors increase monotonically along the prediction window. In this case, $\bbepsilon_i$ could take increasing values based on prior knowledge of the learning problem. We provide further discussions about such configurations in Appendix~\ref{app-monotonic-form}.

Nonetheless, which loss patterns are desirable and attainable will depend ultimately on the model, data and task at hand. In the next section we explore how to \emph{automatically} adapt constraints during training, so that the problem is more robust to their mis-specification.
%
\subsection{Adapting Constraints: Resilient Constrained Learning}\label{subsec:Resilient}

In the unconstrained risk minimization problem ~\eqref{Avg-SRM} an optimal function $\theta^{\star}$ always exists. This is not the case of the constrained learning problem in \eqref{P-LSC} in which there may be no parameters in $\Theta$ that satisfy all the requirements if the target losses $\epsilon_i$ are too restrictive. 

In practice, landing on satisfiable loss shaping requirements may necessitate the relaxation of some constraints --i.e., the values of $\bbepsilon_i$ in~\eqref{P-LSC}--. This is challenging because assessing the impact of tightening or relaxing a particular constraint can have intricate dependencies with the model class, the unknown data distribution, and learning algorithm, which are hard to determine a priori. 

Thus, the point at issue is coming up with reasonable constraints that achieve a desirable trade-off between controlling the distribution of the error across the predictive window and attaining good average performance. That is, for larger constraint levels the average performance improves, although the constraints will have less impact on the optimal function $f_{\theta^{\star}}$. 

 In order to do so, we introduce a non negative perturbation $\bbzeta_t\in\mathbb{R}_+$ associated to each time step, and consider relaxing the $i-th$ constraint in the original problem by $\bbzeta_i$. Explicitly, we impose $\mathbb{E}_{(\x_{c}, y_{p}) \sim \mathfrak{D}}\left[\ell_i({f_{\bbtheta}(\x_{c})}, y_{p})\right] \leq \bbepsilon_i + \bbzeta_i$. We also introduce a differentiable, convex, non-decreasing cost $h: \mathbb{R}_+^{T_{p}} \to \mathbb{R}_+$, that penalizes deviating from the original specification, e.g., the squared $L_2$ norm $ h(\bbzeta) \propto \|\bbzeta\|_2^2$.Thus, we seek a relaxation $\bbzeta^{\star}$ that achieves a desirable trade-off by equating marginal decrease in the objective and the marginal increase in relaxation cost. Explicitly, 

\begin{tcolorbox}
\vspace{-0.15in}
\hspace{-0.15in}
\begin{align}~\label{eq:res} - \partial h(\bbzeta^{\star}) \in \partial P^{\star}(\bbepsilon+\bbzeta^{\star}) ,
\end{align}    
\end{tcolorbox}

 where $\partial P^{\star}$ denotes the subdifferential of $P^{\star}$.
 
 This allows us to re-interpret the initial constraint levels $\bbepsilon$ as a soft constraint, and learn a relaxation that makes the problem easier to solve. As shown by~\cite{hounie2023resilient} this relaxation can be found while jointly solving the learning task by solving the problem
\begin{align*}
\label{Res-LSC}
 &\min_{\bbzeta \in \mathbb{R}^{T_p}}  P^{\star}(\bbepsilon+\bbzeta)+h(\bbzeta)\;.\tag{R-LS}
\end{align*}
%
Further discussion on equivalent formulations that provide  a straightforward way to compute the resilient relaxation are discussed in Appendix~\ref{app:equivalent}.
 In the next section, we introduce practical primal-dual algorithms that enable approximating the constrained~\eqref{P-LSC} and resilient~\eqref{Res-LSC} problems based on samples and finite (possibly non-convex) parametrizations such as transformer architectures.

\section{Empirical Dual Resilient and Constrained Learning}\label{sec:algos}
A challenge in solving~\eqref{P-LSC} and~\eqref{Res-LSC} is that in general, (i) no closed form projection into the feasible set or proximal operator exists, and (ii) it involves an unknown data distribution $\mathfrak{D}$. In what follows, we describe our approach for tackling the Resilient problem, noting that Primal-Dual algorithms tackling the original constrained problem can be similarly derived by simply excluding the slack variables.

To undertake~\ref{Res-LSC}, (i) we replace expectations by sample means over a data set $\{(\boldsymbol{x}_{c}^n, y_{p}^n) \: : \: n=1,\cdots,N \}$, as typically done in (unconstrained) statistical learning, and (ii) resort to its Lagrangian dual.

These modifications lead to the Empirical Dual problem 
\begin{align}\tag{ED-LS}~\label{ED-LSC}
\hat{D}^{\star} = & \max_{\bblambda \geq 0 } \;\; \min_{\boldsymbol{\bbtheta} \in \Theta,\; \bbzeta \in \mathbb{R}^{\tiny T_p}_+} \hat{L}(\boldsymbol{\bbtheta}, \bblambda, \mathbf{\bbepsilon}, \bbzeta),
\end{align}
where $\hat{L}$ is the empirical Lagrangian of~\eqref{Res-LSC}, defined as
\begin{align*}
{\hfilneg \hat{L}(  \boldsymbol{\bbtheta},  \bblambda, \bbepsilon, \bbzeta) := \frac{1}{2} h(\bbzeta)+\hspace{10000pt minus 1fil}}
\end{align*}
\vspace{-0.3in}
\begin{align*}
\frac{1}{N} \sum_{n=1}^N  \sum_{i=1}^{T_{p}} \left( \bblambda_i + \frac{1}{T_{p}} \right) \left[\ell_i({[f_{\bbtheta}(\x_{c}^n)]}, y_{p}^n)\right]- \bblambda_i (\bbepsilon_i+\bbzeta_i), 
\end{align*}
and $\bblambda_i$ is the dual variable associated to the constraint on the $i$-th time step loss $\ell_i$.

The Empirical Dual problem~\eqref{ED-LSC} is an approximation of the Dual problem associated to~\ref{Res-LSC} based on training samples. 
%
%
The dual problem itself can be interpreted as finding the tightest lower bound on the primal. Although the estimation of expectations using sample means and non-convexity of the hypothesis class can introduce a duality gap, i.e., $\hat{D}^{\star}<P^{\star}$, under certain conditions this gap can be bounded.

Unlike unconstrained statistical learning bounds, these approximation bounds depend not only in the sample complexity of the model class and loss but also on the optimal dual variables or slacks associated to the constraint problem. These  reflect how challenging it is to meet the constraints. This aspect is crucial, as applying overly stringent loss shaping constraints in a given learning setting may be detrimental in terms of approximation and lead to sub-optimal test performance. We include a summary of these results as well as a discussion on their implications in this setting in Appendix~\ref{app:bounds}, and refer to~\cite{chamon2020NIPS, hounie2023resilient} for further details. 

The advantage of tackling the the empirical dual problem $\hat{D}^{\star}$ is that it can be solved using saddle point methods presented in the next section.

\subsection{Algorithm}

In order to solve problem~\eqref{ED-LSC}, we resort to dual ascent methods, which can be shown to converge even if the inner minimization problem is non-convex~\cite{chamon2022TIT}. The saddle point problem~\eqref{ED-LSC} can then be undertaken by alternating the minimization with respect to $\bbtheta$ and $\bbzeta$ with the maximization with respect to $\bblambda$~\cite{arrowhurwitz}, which leads to the Primal-Dual constrained learning procedure in Algorithm~\ref{alg:pd}. 

Although a bounded empirical duality gap \emph{does not} guarantee that the primal variables obtained after running Algorithm~\ref{alg:pd} are near optimal or approximately feasible \emph{in general}, recent constrained learning literature provides sub-optimality and near-feasibility bounds for primal iterates~\cite{2024nearoptimal} as well as abundant empirical evidence~\cite{robey2021adversarial, 2022loveconstraints, elenter_active2022} that good solutions can still be obtained.

\begin{algorithm}[h!]
\caption{\textbf{P}rimal \textbf{D}ual \textbf{L}oss \textbf{S}haping.}
\label{alg:pd}
\begin{algorithmic}
\STATE {\bfseries Input:} Dataset $\{x_i, y_i \}_{i=1,\cdots,N} $, primal learning rate $\eta_p$, dual learning rate $\eta_d$, perturbation learning rate $\eta_\epsilon$, number of epochs $T_e$, number of batches $T_b$, initial constraint tightness $\epsilon_{\alpha_0}$.
\STATE Initialize: $\boldsymbol{\theta}, \lambda_1, \dots, \lambda_{T_{\text{pred}}}\leftarrow\mathbf{0}$
\FOR{$\text{epoch } =1, \ldots, T_e \: $}
    \FOR{$\text{batch } =1, \ldots, T_b \: $}
        \STATE \hspace*{5mm}\textit{Update primal variables}
        \STATE $\boldsymbol{\theta} \leftarrow \boldsymbol{\theta} - \eta_p \nabla_{\theta}\hat{L}(\boldsymbol{\theta}, \boldsymbol{\lambda}, \bbepsilon, \bbzeta)$
        \STATE \hspace*{5mm}\textit{Evaluate constraints.}
        \STATE $\mathbf{s}_i \leftarrow \left( \frac{1}{N_b} \sum_{n=1}^{N_b}\ell_i({[f_{\boldsymbol{\theta}}(\mathbf{x}_{\text{c}(t)}^n)]}, y_{\text{p}(t)}^n)  \right) - (\epsilon_i+\bbzeta_i)$
        \STATE \hspace*{5mm}\textit{Update slacks.}
        \STATE $\bbzeta \leftarrow \left[ \bbzeta - \eta_\zeta \Big( \nabla h\big( \bbzeta \big) - \bblambda \Big)\right]_{+}$
        \STATE \hspace*{5mm}\textit{Update dual variables.}
        \STATE $\boldsymbol{\lambda} \leftarrow \left[\lambda + \eta_d \mathbf{s} \right]_{+}$
    \ENDFOR
\ENDFOR
\STATE {\bfseries Return:}{  $\boldsymbol{\theta}$, $\boldsymbol{\lambda}$,
$\bbzeta$.}
\end{algorithmic}
\end{algorithm}

\begin{figure*}[t]
        \centering
         \includegraphics[width=0.99\linewidth, trim={0 0 .1in 0}, clip]{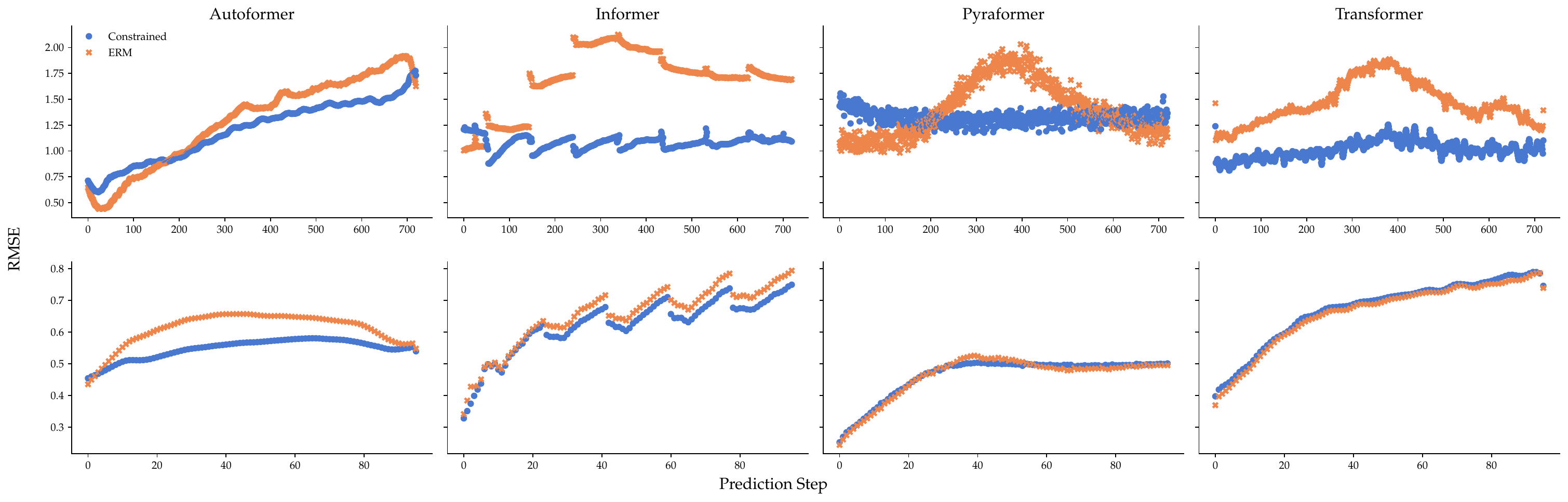}
    \caption{Test MSE at each prediction step for two datasets across different models, datasets and predictive windows. The top row shows results for the Weather dataset with a predictive window length of 96 steps, and the second row corresponds to the Exchange Rate dataset with a predictive window length of 720 steps. 
    Each column corresponds to a different architecture, and each curve represents a different training algorithm, we include both the ERM baseline and our method using a constant constraint across the window for all models.
    }
    \label{fig:constr_vs_erm}
\end{figure*}

\section{Experiments}\label{sec:exps}
We conduct extensive evaluations comparing constrained and resilient constrained learning against the customary unconstrained training pipeline across eight model architectures and nine popular datasets. For each dataset, we train models with four different predictive window lengths to evaluate the impact of constraints as the forecasting horizons extended. In total, this amounts to 288 different experiment settings.

Explicitly, the datasets are: Electricity Consumption Load (ECL), Weather, Exchange Rate~\cite{2018ExchangeLSTN}, Traffic, Electricity Transformer Temperature (ETT) (two hourly datasets and two every 15 minutes)~\cite{2021informer}, and Influenza Like Illness (ILI). Following the literature, we train with predictive window lengths of 96, 192, 320, and 720 for each dataset, except for Illness, which is trained to predict with lengths of 24, 36, 48, and 60. For a more detailed explanation of the datasets' contents and source, refer to the Appendix~\ref{app-datasets}.

We also include a wide variety of time series prediction models comprising seven transformer-based architectures and one non-transformer architecture. Namely, the transformer models are: Reformer~\cite{kitaev2019reformer}, Autoformer~\cite{wu2021autoformer}, Informer~\cite{2021informer}, Pyraformer~\cite{2021pyraformer}, iTransformer~\cite{liu2024itransformer}, Nonstationary Transformer~\cite{liu2022nonstationary} and a vanilla transformer architecture~\cite{2017VaswaniAttention}. The non-transformer model is FiLM~\cite{2022FiLM}. 

We follow the same setup, including preprocessing, hyperparameters and implementation, as described in~\cite{kitaev2019reformer, wu2021autoformer, 2021pyraformer,  2021informer,liu2022nonstationary, 2022FiLM}. For our method, we only perform a grid search over six values of the constraint level $\epsilon$. The constrained and resilient loss shaping results presented in this section correspond to the best performing constraint level. We provide an additional ablation analysis on one dataset in Appendix~\ref{app-ablation}. 

Data is split into train, validation, and test chronologically with a ratio of 7:1:2. For each data split we extract every  pair of consecutive context and prediction windows of length $T_{c}$ and $T_{p}$. That is, we use rolling (overlapping) windows for both training and testing.
\footnote{We also fix a known bug~(reported in \cite{google-bug}) from~\cite{wu2021autoformer} and followup works, where the last $T_{p}$ samples where discarded.} 
Additional experiment details can be found on Appendix~\ref{app-exp-setup}.
\subsection{Loss Shaping}\label{subsec:loss-reshaping}

In this section, we demonstrate that our approach effectively reduces performance fluctuations across the window while maintaining comparable average performance. We compute the mean squared errors (MSE) at each predicted timestep and report both the average MSE and the standard deviation across the window, referred to as Window STD.

To ease interpretation and facilitate comparisons across different experimental settings, we normalize errors using the unconstrained ERM baseline. For complete result tables with unnormalized MSE values and additional example cases, refer to Appendix~\ref{app-exp-results}.

\begin{figure*}[t]
    \centering

    \includegraphics[width=0.45\textwidth]{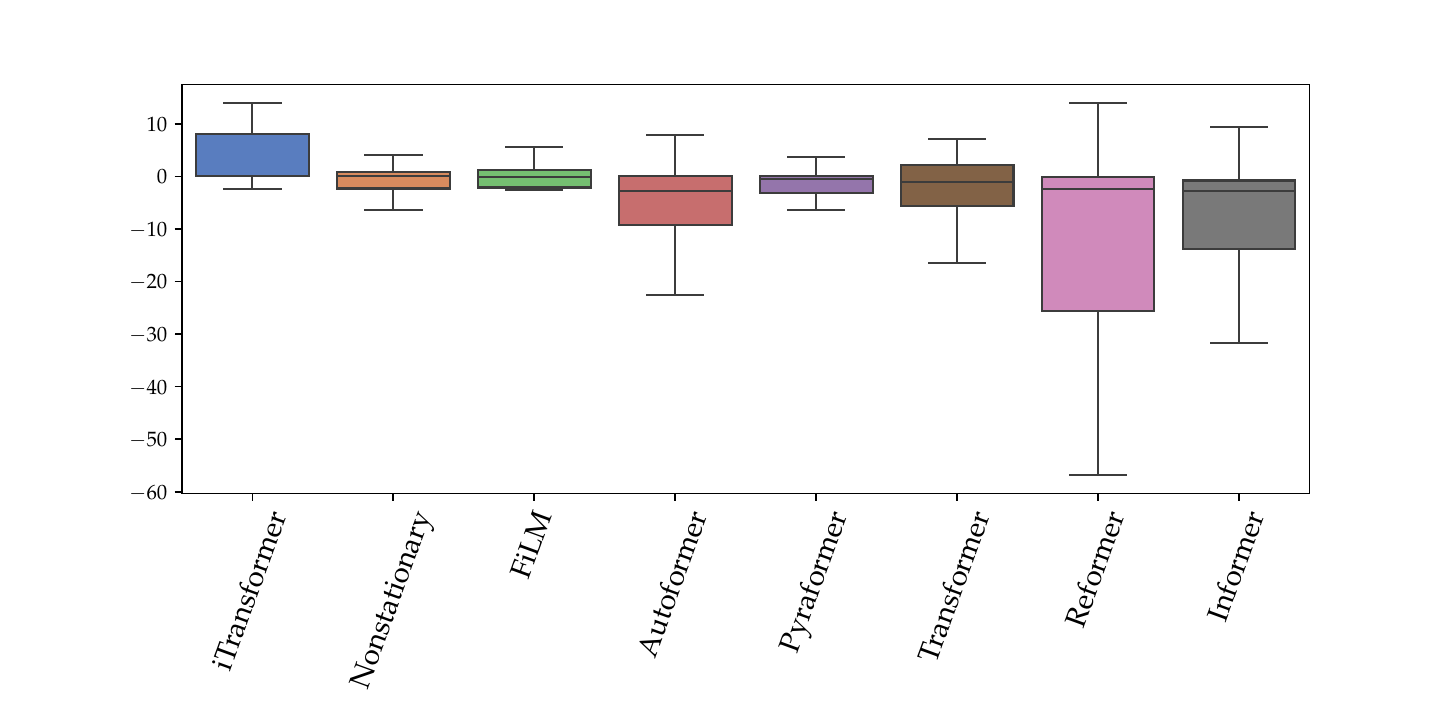}
    \includegraphics[width=0.45\textwidth]{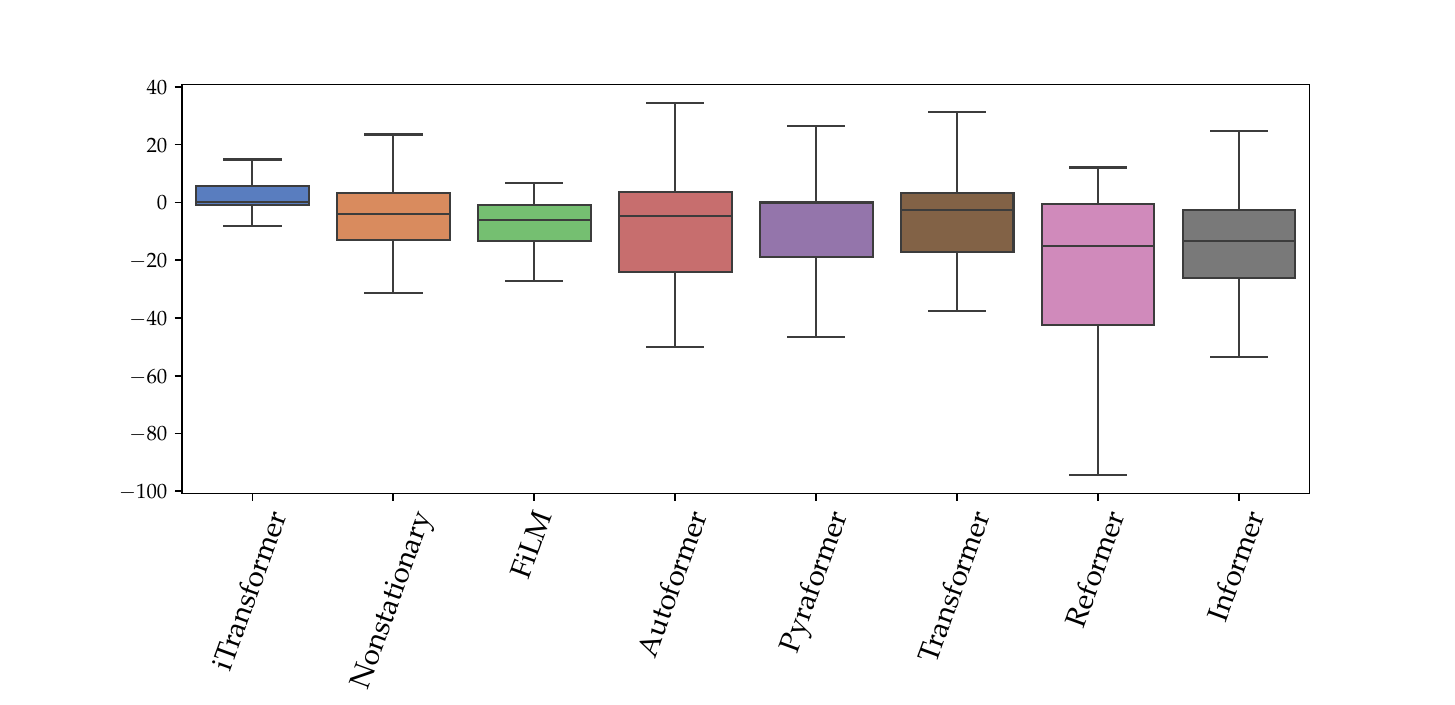}
    \includegraphics[width=0.45\textwidth]{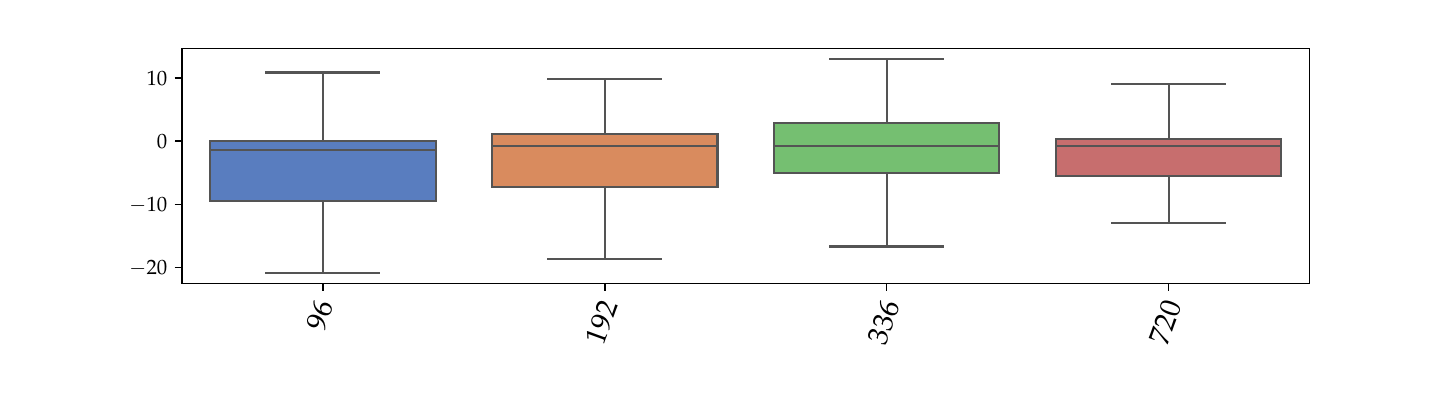}
    \includegraphics[width=0.45\textwidth]{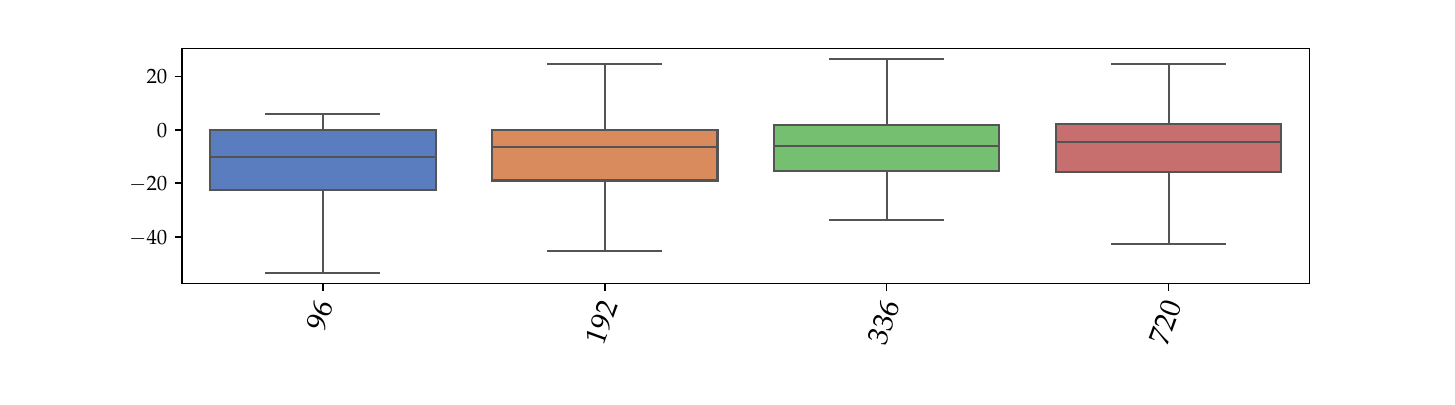}
    \caption{Box plots of percentual changes across all experiments. The left column contains plots of MSE across experiments, and the right column is Window STD. We segment experiments by model and prediction length. The $x$ axes of each box plot are sorted by the mean ERM MSE (better models and easier datasets to the left). The full table with the results for each ERM, constrained and resilient setting can be found in Appendix \ref{app-exp-results}.}
    \label{fig:boxplots}
\end{figure*}

We do not to use constraint violation as a metric because it can be uninformative when the loss landscape is consistently infeasible due to a large generalization gap and constraint levels are constant. In such cases, constraint violation is as informative as MSE, even when the error distribution  over time changes.

\begin{figure*}[t]
    \centering
    \includegraphics[width=0.43\textwidth]{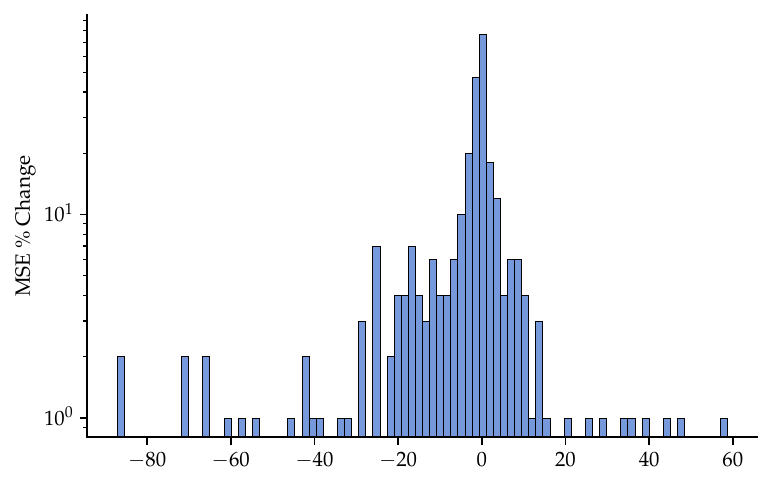}
    \includegraphics[width=0.43\textwidth]{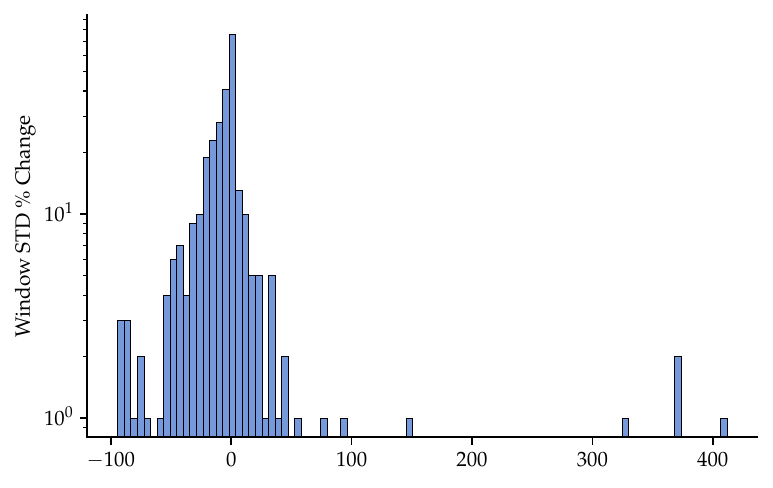}
    \caption{Distribution of percentual change in test MSE and Window STD across experiment settings when comparing a run of ERM with a constrained run. On average, STD decreases by 3.47\% and MSE by 4.47\% respectively.}
    \label{fig:histograms}
\end{figure*}

Due to the large number of experimental setups, we summarize quantitative results in Figure~\ref{fig:histograms}, which illustrates the relative differences in MSE and Window STD between ERM and constrained runs. It shows that loss shaping constrained models typically achieve reductions in Window STD while preserving or even improving average MSE. 

We then analyze how the performance of our method varies across setups, by grouping results by prediction length, model architecture and dataset. First, it varies significantly across models, as shown in the second row of Figure~\ref{fig:boxplots}. We conjecture that this variability can be attributed to the inductive biases of certain architectures, which makes it harder to impose the desired loss landscape. Second, the relative effect of our method on both MSE and Window STD remains largely unchanged as the prediction window increases, as shown in the third row of box plots in Figure~\ref{fig:boxplots}. Since both average MSE and Window STD increase with prediction length, a constant relative effect means that the absolute change  \textit{increases} with prediction length. Appendix~\ref{app-loss-shapes} shows several concrete examples of this phenomenon.

To qualitatively assess the impact of our method on the error distribution, we include plots of per time-step losses comparing models trained with ERM and our method in Figure~\ref{fig:constr_vs_erm}. In these settings, our approach effectively affects the distribution of losses across the prediction windows. For example, in the Weather dataset (first row), Autoformer and Pyraformer are trading off overall MSE for a flatter error across the window. In other cases, like the Exchange Rate dataset for the Informer and Transformer models, in addition to showing flatter landscapes, the constrained models perform better than their ERM counterparts overall. 

\begin{figure}[h]
        \centering
        \includegraphics[width=0.95\linewidth,  trim={0 0.1in 0 0}, clip]{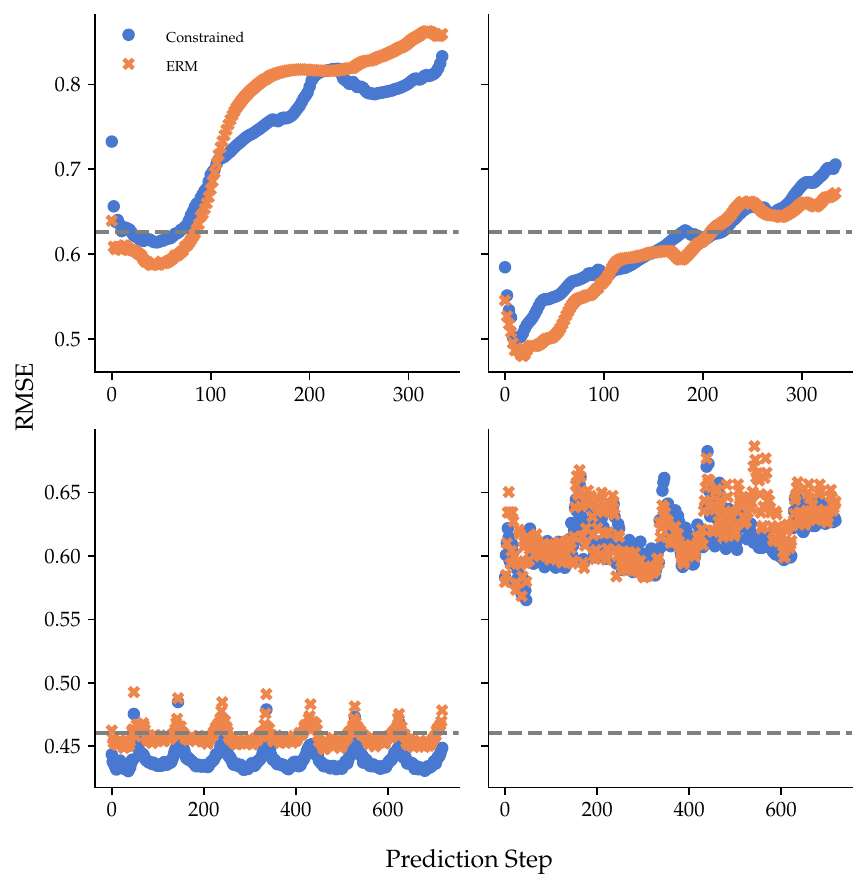}
    \caption{Two instances of failure cases. The first row is the training and testing errors of Autoformer on Weather data with window length of 336. The second row is Informer on ECL data with window length of 720. The gray lines are the values of $\bbepsilon$ used during training.}\label{fig:failures}
\end{figure}

While we observe that in many cases imposing constraints is beneficial, we also distinguish two failure modes where we are unable to change test loss as desired. The first is when constraints are not feasible at the end of training, as seen in the first row of Figure~\ref{fig:failures}. This is the motivation behind the resilient approach, presented in the next section. A second failure mode is due to the inherent generalization gap, which results in feasible constraints during training, but no effective loss shaping in test data, such as in the second row of Figure~\ref{fig:failures}.  We defer to Appendix~\ref{app-loss-shapes} for more in-depth comparisons of ERM and constrained runs.

\begin{figure}[!h]
        \centering
        \includegraphics[width=0.97\linewidth,  trim={0 0.11in 0 0}, clip]{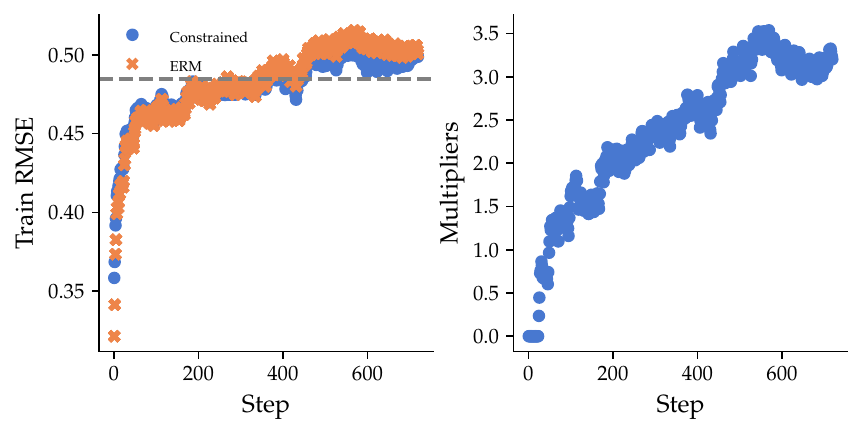}
    \caption{
    Training MSE and multipliers of a constrained model, with the ERM curve for reference. The setting is an Autoformer model for the Exchange Rate dataset, with a predictive window length of 720 steps. The gray line is the constraint, with a value of $\sqrt{\epsilon}=0.484$.
    }
    \label{fig:constrained_train}
\end{figure}

Another advantage of our primal-dual approach is that multipliers indicate the difficulty of satisfying each constraint. Therefore, we can examine the impact of constraints on loss landscapes on a run by analyzing the loss distributions and multipliers in Figure~\ref{fig:constrained_train}. In this experiment, the training loss increases approximately monotonically over the prediction window. The final solution is infeasible, as the MSE constraints are violated beginning at step 341. Consequently, the optimal multipliers are also high, reflecting the difficulty of satisfying the constraints in the latter part of the window. Despite the growth of the multipliers, the model still fails to satisfy the constraints. This infeasibility during training results in ineffective loss shaping during testing.

\subsection{Resilience}

\begin{figure}[h]
        \centering
        \includegraphics[width=0.97\linewidth,  trim={0 0.11in 0 0}, clip]{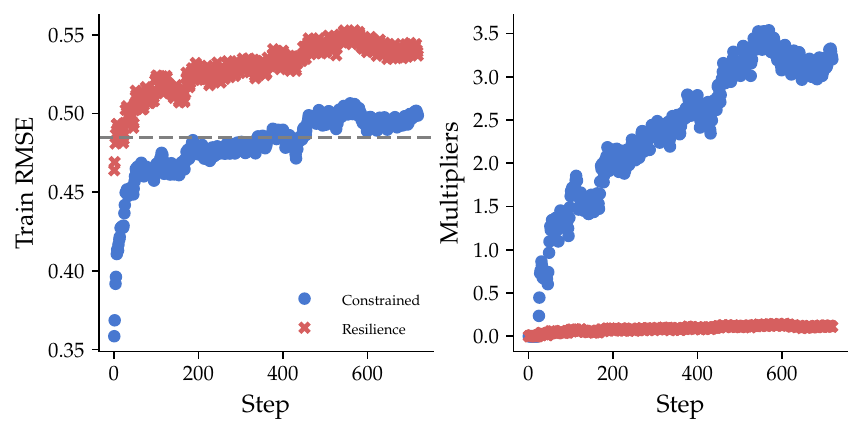}
    \caption{
        Training MSE and multipliers of constrained and resilient models. Resilience finds a feasible solution by relaxing the harder constraints. The setting is the same as in Figure ~\ref{fig:constrained_train}.
    }
    \label{fig:resilience_train}
\end{figure}

Resilient constrained learning can effectively overcome feasibility issues in training --like those mentioned in the previous section-- by relaxing  the most difficult constraints. As shown in Figure~\ref{fig:resilience_train}, when comparing hard constraints with resilient constraints using the same setting as the previous section, the resilient approach not only leads to smaller multipliers by relaxing some constraints, but also yields a solution that is more feasible overall.

\begin{figure}[h]
        \centering
        \includegraphics[width=0.97\linewidth,  trim={0 0.11in 0 0}, clip]{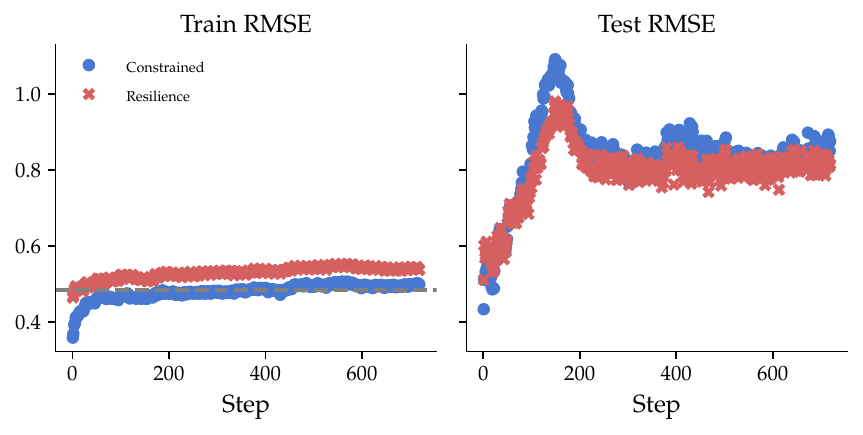}
    \caption{
    Train and test MSE of constrained and resilient models. By finding a feasible solution in training, generalization improves. The setting is the same as in Figure~\ref{fig:constrained_train}.
    }
    \label{fig:resilience_generalization}
\end{figure}

In addition, we find that the solutions achieved by the resilient approach can also have better generalization when compared to hard constraints, as seen in Figure~\ref{fig:resilience_generalization}. The resilient model reduces the peak around step 150. This illustrates how resilient learning effectively mitigates the failure mode mentioned in the previous section.

As a result of relaxing constraints and solving feasibility issues during training, resilient models are effective in reducing Window STD in a wide variety of settings. This is supported by our empirical results in Tables~\ref{tab:window_std_pt1} and~\ref{tab:window_std_pt2}. For instance, resilient FiLM achieves the lowest Window STD when compared to constrained and ERM-trained models. Models such as Autoformer, Pyraformer and iTransformer also achieve its lowest Window STD when using Resilient constraints. 

Furthermore, we find that some settings also achieve its lowest MSE when using resilient constraints. Such is the case of iTransformer on datasets ETTh2,ETTm1 and ETTm2, where the resilient model has the best overall error in almost all predictive window lengths, as shown in the full MSE results, Tables~\ref{tab:mse_all_pt1} and ~\ref{tab:mse_all_pt2} in Appendix~\ref{app-exp-results}. 

\begin{table*}[b]
\centering
\begin{tabular}{cc|ccc|ccc|ccc}
\hline
                                       &     & \multicolumn{3}{c|}{Nonstationary Transformer} & \multicolumn{3}{c|}{Pyraformer}   & \multicolumn{3}{c|}{Reformer}                                          \\
                                       &     & ERM        & Constant     & Exp                & ERM   & Constant & Exp            & ERM            & Constant       & \multicolumn{1}{c|}{Exp}           \\ \hline
\multirow[t]{4}{*}{\multirow{4}{*}{\rotatebox{90}{ECL}}} & 96  & 0.178      & 0.182        & \textbf{0.170}     & 0.287 & 0.278    & \textbf{0.274} & 0.299          & \textbf{0.297} & \multicolumn{1}{c|}{0.298}          \\
                                       & 192 & 0.186      & 0.191        & \textbf{0.180}     & 0.294 & 0.292    & \textbf{0.289} & 0.332          & 0.326          & \multicolumn{1}{c|}{\textbf{0.324}} \\
                                       & 336 & 0.204      & 0.204        & \textbf{0.196}     & 0.313 & 0.302    & \textbf{0.295} & \textbf{0.359} & 0.365          & \multicolumn{1}{c|}{0.364}          \\
                                       & 720 & 0.233      & 0.229        & \textbf{0.228}     & 0.310 & 0.298    & \textbf{0.296} & \textbf{0.316} & 0.316          & \multicolumn{1}{c|}{0.318}          \\ \hline
\end{tabular}
\caption{Test MSE of ERM, constant-constrained and exponentially-constrained settings in the ECL dataset, for three model architectures}
\label{tab:exponential_mse}
\end{table*}
\begin{table*}
    \centering
\begin{tabular}{cc|ccc|ccc|ccc}
\hline
                                       &     & \multicolumn{3}{c|}{Nonstationary Transformer} & \multicolumn{3}{c|}{Pyraformer}                  & \multicolumn{3}{c|}{Reformer}                                          \\
                                       &     & ERM         & Constant            & Exp        & ERM            & Constant          & Exp            & ERM            & Constant          & \multicolumn{1}{c|}{Exp}            \\ \hline
\multirow[t]{4}{*}{\multirow{4}{*}{\rotatebox{90}{ECL}}} & 96  & 0.017       & \textbf{0.014}      & 0.016      & 0.019          & \textbf{0.015} & \textbf{0.015} & \textbf{0.011} & 0.011          & \multicolumn{1}{c|}{0.012}          \\
                                       & 192 & 0.026       & \textbf{0.019}      & 0.020      & 0.022          & \textbf{0.017} & 0.022          & 0.020          & 0.020          & \multicolumn{1}{c|}{\textbf{0.018}} \\
                                       & 336 & 0.031       & \textbf{0.024}      & 0.025      & 0.034          & \textbf{0.023} & 0.028          & \textbf{0.022} & 0.023          & \multicolumn{1}{c|}{0.023}          \\
                                       & 720 & 0.040       & \textbf{0.037}      & 0.038      & \textbf{0.015} & 0.016          & 0.016          & \textbf{0.007} & \textbf{0.007} & \multicolumn{1}{c|}{\textbf{0.007}} \\ \hline
\end{tabular}
\caption{Window STD of ERM, constant-constrained and exponentially-constrained settings in the ECL dataset, for three model architectures.}
\label{tab:exponential_std}
\end{table*}


\subsection{Monotonically Increasing Constraints}

The results of the previous sections all use $\bbepsilon_i$ set to a constant value derived from the statistics of training or validation ERM errors. As discussed in Section~\ref{sec:Constrained}, an alternative is to have the constraints take monotonically increasing values, for instance, from a linear or exponential fit. Having increasingly looser constraints is only logical in the common scenario where errors are expected to grow as we predict farther into the future. Note that we call these constraints exponential because they were set from an exponential fit, but they are linear in the optimization variable.

An empirical evaluation on the ECL dataset with three different models shows this is a viable constraint design. Table~\ref{tab:exponential_mse} contains the MSE for each of the three models: in two cases, exponential achieves the lowest MSE across all windows. With respect to Window STD, Table~\ref{tab:exponential_std} shows that exponential does not always find the most stable shape. However, this is expected, because the monotonically increasing constraint will not necessarily enforce a reduction in variance across the window.

\section{Conclusion}\label{sec:conclu}

This paper introduced a time series forecasting constrained learning framework that aims to find the best model in terms of average performance while imposing a user-defined upper bound on the loss at each time-step. Given that we observed that the distribution of loss landscapes varies considerably, we explored a resilient constrained learning approach to dynamically adapt individual constraints during training. We analyzed the properties of this problem by leveraging recent duality results and developed practical algorithms to tackle it. We empirically corroborated that our approach can effectively alter the loss distribution across the forecast window.

Although we have focused on transformers for long-term time series forecasting, motivated by the empirical finding that their loss varies considerably, the loss shaping constrained learning framework can be extended to other settings. This includes a plethora of other models, datasets and tasks where losses are distributed across different time steps or grids. Furthermore, in this work, we focused on constraints characterized by a constant value across all timesteps, but other types of constraints, such as exponentially increasing stepwise constraints that we briefly presented, could also be explored in the future.

\newpage

\section*{Impact Statement}
This paper presents work whose goal is to advance the field of Machine Learning. There are many potential societal consequences of our work, none which we feel must be specifically highlighted here.

\bibliographystyle{icml2024}
\bibliography{refs}

\newpage
\appendix

\section{Approximation Guarantees}~\label{app:bounds}
The functional space induced by the parametrization, which we will denote $\calF_{\theta} = \{ f_\theta: \bbtheta \in \Theta \}$ can be non-convex, as in the case of transformer architectures. However, it has been shown~\cite{chamon2020NIPS, hounie2023resilient} that as long as the distance to its convex hull $\overline{\calF}_{\theta} = \conv({\calF}_{\theta}) $ is bounded, then we can leverage the strong duality of the convex variational program defined over $\overline{\calF}_{\theta}$ together with uniform convergence bounds to provide approximation guarantees. That is, the values of~\eqref{Res-LSC} and~\eqref{ED-LSC} are close. This holds both for the constrained and resilient formulation, although as discussed next, the resilient problem requires milder assumptions.

\begin{assumption}\label{ass_constraint_qualification}
There exist a function~$\phi \in \calF$ such that all constraints are met with margin~$c > 0$, i.e.,
$\mbE_{\calD_i} \big[ \ell_i( \phi(\bbx),y ) \big] \leq  c$, for all~$i = 1, \dots, m$.
\end{assumption}
\vspace{-.1in}
\textbf{Assumption A.1${.^{\prime}}$}\label{ass_constraint_qualification_constrained}
There exist a finite relaxation~$\bbzeta \preceq\infty$ and a function~$\phi \in \calF$ such that all constraints are met with margin~$c > 0$, i.e.,
$\mbE_{\calD_i} \big[ \ell_i( \phi(\bbx),y ) \big] \leq \bbzeta - c$, for all~$i = 1, \dots, m$.
\begin{assumption}\label{ass_losses_Lipchitz}
The loss functions~$\ell_i$, $i=0 \dots m$, are convex and~$M$-Lipschitz continuous.
\end{assumption}
\begin{assumption}\label{ass_parametrization}
For every~$\phi \in \calF$, there exists~$\bbtheta^\dagger \in \Theta$ such that
$\mbE_{\calD_i} \big[ \vert \phi(\bbx) - f_{\bbtheta^\dagger}(\bbx) \vert \big]\leq \nu$, for all~$i = 0,\dots,m$.
\end{assumption}

\begin{assumption}\label{ass_uni_conv}
There exists~$\xi(N, \delta) \geq 0$ such that for all~$i = 0,\dots,m$ and all~$\bbtheta \in \Theta$,
\begin{equation*}
	\bigg\vert
		\mbE_{\calD_i} \big[ \ell_i( f_{\theta}(\bbx_c),y_p ) \big]
		- \frac{1}{N} \sum_{n=1}^{N} \ell_i\big( f_{\theta}(\bbx_c^{n}),y_p^{n} \big)
	\bigg\vert \leq \xi(N, \delta)
\end{equation*}
with probability~$1-\delta$ over draws of~$\{(\bbx_{i}^n, y_{i}^n)\}$.
\end{assumption}

Note that the constraint qualification in Assumption~\ref{ass_constraint_qualification} (known as Slater's condition~\cite{boyd2004convex}), which is required for the constrained problem, can be relaxed to the milder qualification given in Assumption A.1${.^{\prime}}$ for the resilient problem.

Assumption~\ref{ass_losses_Lipchitz} holds for commonly used objectives in Time-Series forecasting, including Mean Squared Error, Mean Absolute Error and Huber-loss, among others.

Assumption~\ref{ass_parametrization} will hold if the parametrized function space is rich in the sense that the distance to its convex hull is bounded. Since neural networks and transformers have universal approximation properties~\cite{MLP-Universal-approx, 2019transformersUniversalApprox}, we posit that given a parametrization with enough capacity, this assumption holds. 

Assumption~\ref{ass_uni_conv} -- known as uniform convergence -- is customary in statistical learning theory. This includes generalization bounds based on VC dimension, Rademacher complexity, or algorithmic stability, among others~\cite{mohri2018foundations-of-ml}. Unlike generalization bounds for the unconstrained problem~\eqref{Avg-SRM}, this assumption bounds the loss at each step, and not the aggregated loss throughout the window.

We now state the main theorems that bound the approximation errors.

\textbf{Theorem 1~\cite{hounie2023resilient}} 
Let~$\bbzeta^\star$ be an optimal relaxation for problem~\eqref{Res-LSC}. Under Ass. A.1$^\prime$--\ref{ass_uni_conv}, it holds with probability of~$1-(3m+2)\delta$ that
\begin{equation*}\label{eqn_main_bound_res}
    \big\vert P^{\star} - D^{\star} \big\vert
    	\leq h(\bbzeta^{\star} + 1 \cdot M\nu) - h(\bbzeta^\star) + M \nu + (1 + \Delta) \xi(N,\delta)
    	\text{.}
\end{equation*}

\textbf{Theorem 2~\cite{chamon2020NIPS}} 
Let~$\bblambda^\star$ be an optimal relaxation for the constrained problem~\eqref{P-LSC} defined over $\overline{\calF}_{\theta}$ with constraints $\bbepsilon_i-M \nu$. Under Ass.~\ref{ass_constraint_qualification}--\ref{ass_uni_conv}, it holds with probability of~$1-(3m+2)\delta$ that
\begin{equation*}\label{eqn_main_bound_res}
    \big\vert P^{\star} - D^{\star} \big\vert
    	\leq (1+\bblambda^\star) M \nu + (1 + \Delta) \xi(N,\delta)
    	\text{.}
\end{equation*}

Note that unlike in unconstrained statistical learning, optimal dual variables $\bblambda^\star$ or relaxations~$\bbzeta^\star$ play an important role in these approximation bounds. This is because, intuitively, they represent the difficulty of satisfying constraints.
\section{Monotonic Loss constraints}~\label{app:monotonic}
 A less restrictive constraint that accounts for error compounding is to only require errors to be monotonically increasing. That is, that there exists some $\bbepsilon$ such that $\bbepsilon_{t+1}\geq\bbepsilon_t$ for all $t$. Instead of searching for such an $\bbepsilon$, we can directly impose monotonicity, which requires only a small modification in ~\eqref{P-LSC}, as discussed in Section~\ref{app-monotonic-form}.

 \subsection{Formulation}\label{app-monotonic-form}
 
 We can require the loss on each timestep $i$ to be smaller than the loss step at the following $i+1$. This leads to the constrained statistical learning problem:
 
\begin{tcolorbox}
\vspace{-0.15in}
\begin{align*}
\label{P-Mono}
& \min_{\bbtheta \in \Theta}  \; \mathbb{E}_{(\x_{c}, y_{p}) \sim \mathfrak{D}}\left[\frac{1}{T_{p}} \sum_{i=1}^{T_{p}} \ell_i({f_{\bbtheta}(\x_{c})}, y_{p}) \right] \\
&\text{s. to:}\; \mathbb{E}_{(\x_{c}, y_{p}) \sim \mathfrak{D}}\left[\ell_i({f_{\bbtheta}(\x_{c})}, y_{p})\right] \nonumber\\ 
& \;\quad\quad\quad\quad  \leq \mathbb{E}_{(\x_{c}, y_{p}) \sim \mathfrak{D}}\left[\ell_{i+1}({f_{\bbtheta}(\x_{c})}, y_{p})\right] \; \nonumber\\
& \quad\quad\quad\quad\quad\quad\quad\quad i=1,\ldots, T_{\text{pred}-1}. \tag{P-M}
\end{align*}
\end{tcolorbox}

Because of linearity of the expectation, the constraint can be re-written as a single expectation of a (non-convex) functional, explicitly,
\begin{align*}
\mathbb{E}_{(\x_{c}, y_{p}) \sim \mathfrak{D}} & \left[ \ell_i({f_{\bbtheta}(\x_{c})}, y_{p}) \right. \\
 & \quad\quad\quad \left. - \ell_{i+1}({f_{\bbtheta}(\x_{c})}, y_{p})\right] \leq  0.
\end{align*}

Despite the lack of convexity, the bounds presented in Appendix~\ref{app:bounds} still hold under additional assumptions on the distribution of the data and model class, as shown in Theorem 1 and Proposition B.1~\cite{chamon2022TIT}. We include these assumptions here for completeness.

\begin{assumption}
     The functions $y \mapsto \ell_i(\phi(\cdot), y) f_i(\cdot \mid y), i=$ $1, \ldots, m$, are uniformly continuous in the total variation topology for each $\phi \in \mathcal{\mathcal { H }}$, where $f_i(\boldsymbol{x} \mid y)$ denotes the density of the conditional random variable induced by $\mathfrak{D}_i$. Explicitly, for each $\phi \in \overline{\mathcal{H}}$ and every $\epsilon>0$ there exists $\delta_{\phi, i}>0$ such that for all $|y-\tilde{y}| \leq \delta_{\phi, i}$ it holds that
$$
\sup _{\mathcal{Z} \in \mathcal{B}} \int_{\mathcal{Z}}\left|\ell_i(\phi(\boldsymbol{x}), y) f_i(\boldsymbol{x} \mid y)-\ell_i(\phi(\boldsymbol{x}), \tilde{y}) f_i(\boldsymbol{x} \mid \tilde{y})\right| d \boldsymbol{x} \leq \epsilon
$$
\end{assumption}

\begin{assumption}
    The conditional distribution $\boldsymbol{x} \mid y$ induced by $\mathfrak{D}$ are non-atomic.
\end{assumption}

Note that the epigraph formulation of~\eqref{P-Mono} resembles that of the original los schaping constrained problem~\eqref{P-LSC}, with the difference that $\bbepsilon$ is an optimization variable, now constrained to be monotonically increasing:
\begin{align*}
\label{P-Mono-epi}
& \min_{\bbtheta \in \Theta}  \; \mathbb{E}_{(\x_{c}, y_{p}) \sim \mathfrak{D}}\left[\frac{1}{T_{p}} \sum_{i=1}^{T_{p}} \ell_i({f_{\bbtheta}(\x_{c})}, y_{p}) \right] \\
&\text{s. to :} \quad \mathbb{E}_{(\x_{c}, y_{p}) \sim \mathfrak{D}}\left[\ell_i({f_{\bbtheta}(\x_{c})}, y_{p})\right] \leq \bbepsilon_i \; \nonumber\\
&  \quad\quad \quad \quad\quad\quad \quad \quad \bbepsilon_i \leq \bbepsilon_{i+1}\nonumber\\
& \quad\quad \quad \quad \quad \quad\quad\quad i=1,\ldots, T_{p}-1.\tag{P-M-epi}
\end{align*}
%
\subsection{Equivalent Resilient formulations}~\label{app:equivalent}
The resilient relaxation, as defined in Equation~\ref{eq:res} can be found by solving the \emph{Resilient} Constrained Learning problem~\ref{Res-LSC}.
Note that evaluating $P^{\star}(\bbepsilon+\bbzeta)$ amounts to solving the a constrained learning problem. As shown by~\cite{hounie2023resilient} the relaxation and model can be found jointly by solving:%
\begin{align*}
\label{Res-LSC}
 &\min_{\bbtheta \in \Theta}  \; \mathbb{E}_{(\x_{c}, y_{p}) \sim \mathfrak{D}}\left[\frac{1}{T_{p}} \sum_{i=1}^{T_{p}} \ell_i({f_{\bbtheta}(\x_{c})}, y_{p}) \right] + h(\bbzeta) \\
&\text{s. to :} \quad \mathbb{E}_{(\x_{c}, y_{p}) \sim \mathfrak{D}}\left[\ell_i({f_{\bbtheta}(\x_{c})}, y_{p})\right] \leq \bbepsilon_i+ \bbzeta_i \; \nonumber\\
& \quad\quad i=1,\ldots, T_{p}. \tag{R-LS}
\end{align*}
%

The Lagrangian associated with this problem is 
\begin{align*}
L(\boldsymbol{\theta}, \boldsymbol{\lambda}, \boldsymbol{\epsilon}, \boldsymbol{\bbzeta}) :=&  h(\boldsymbol{\bbzeta}) \\
&+ \sum_{i=1}^{T_{p}} \left( \bblambda_i + \frac{1}{T_{p}} \right) \\
&\quad \mathbb{E}_{(\mathbf{x}_{c}, y_{p}) \sim \mathfrak{D}} 
\left[\ell_i \left(f_{\boldsymbol{\theta}}(\mathbf{x}_{c}), y_{p}\right)\right] \\
&- \bblambda_i (\epsilon_i + \bbzeta_i).
\end{align*}

Since it is separable in $\bbtheta$ and $\bbzeta$, we can re-write the joint minimization as
\begin{align}~\label{eq:decomp}
   \min_{\bbzeta \in \mathbb{R}^{T_p}, \bbtheta \in \Theta} L(  \boldsymbol{\bbtheta},  \bblambda, \bbepsilon, \bbzeta) &=  \min_{\bbzeta \in \mathbb{R}^{T_p}} L_{\bbzeta}(\bblambda, \bbzeta) \\
   &+ \min_{ \bbtheta \in \Theta} L_{\bbtheta}( \boldsymbol{\bbtheta},  \bblambda, \bbepsilon) , \nonumber
\end{align}
where 
\begin{align*}
L_{\bbzeta}(\bblambda, \bbzeta) = h(\bbzeta) - \bblambda^T \bbzeta, \end{align*} 
and 
\begin{align*}
L_{\bbtheta}( \boldsymbol{\bbtheta},  \bblambda, \bbepsilon) = & \sum_{i=1}^{T_{p}} \left( \bblambda_i + \frac{1}{T_{p}} \right)
\mathbb{E}_{(\mathbf{x}_{c}, y_{p}) \sim \mathfrak{D}} 
\left[\ell_i \left(f_{\boldsymbol{\bbtheta}}(\mathbf{x}_{c}), y_{p}\right)\right]\\
& - \bblambda^T \bbepsilon.
\end{align*}
Since the minimization over $\bbzeta$ in Equation~\ref{eq:decomp} yields $-h^*(\bblambda)$, where $-h^*$ is the convex conjugate of $h$, we can re-write the Lagrangian minimization as 
\begin{align*}
     \min_{\bbzeta \in \mathbb{R}^{T_p}, \bbtheta \in \Theta} L(  \boldsymbol{\bbtheta},  \bblambda, \bbepsilon, \bbzeta) = \min_{ \bbtheta \in \Theta} L_{\bbtheta}( \boldsymbol{\bbtheta},  \bblambda, \bbepsilon) - h^*({\bblambda}).
\end{align*}
 Then the dual problem associated to~\ref{Res-LSC} is equivalent to
\begin{align*}
D^{\star} = & \max_{\bblambda \geq 0 } \;\; \min_{\boldsymbol{\bbtheta} \in \Theta} L_{\bbtheta}( \boldsymbol{\bbtheta},  \bblambda, \bbepsilon) - h^*({\bblambda}),
\end{align*}
which corresponds to the original dual problem with a regularization on dual variables given by $-h^*(\bblambda)$.

Although this equivalent formulation was not used directly in this work, which instead solves for $\bbzeta$ using gradient descent as described in the next section, it helps to interpret how the resilient approach makes the problem easier to solve by regularizing the dual function.

\subsection{Algorithm development}

Algorithm~\ref{alg:pd} aims to find a saddle-point of the empirical Lagrangian 
\begin{align*}
{\hfilneg \hat{L}(  \boldsymbol{\bbtheta},  \bblambda, \bbepsilon, \bbzeta) := h(\bbzeta)+\hspace{10000pt minus 1fil}}
\end{align*}
\vspace{-0.3in}
\begin{align*}
\frac{1}{N} \sum_{n=1}^N  \sum_{i=1}^{T_{p}} \left( \bblambda_i + \frac{1}{T_{p}} \right) \left[\ell_i({[f_{\bbtheta}(\x_{c}^n)]}, y_{p}^n)\right]- \bblambda_i (\bbepsilon_i+\bbzeta_i), 
\end{align*}
The updates in Algorithm 1 use gradient descent for $\bbzeta$ and stochastic sub-gradient ascent for $\bblambda$. However, nothing precludes our method from using other optimization algorithms.

To obtain the gradients of the Lagrangian with respect to the primal variables, we exploit the fact that the minimization of the Lagrangian can be separated into two parts, each depending on only one primal variable, as described in the previous section.

The gradient with respect to $\bbzeta$ is 
\begin{align*}
  d\bbzeta = \frac{\partial L(\bblambda,\bbzeta)}{\partial \bbzeta}=\nabla h(\bbzeta) - \bblambda.  
\end{align*}

For dual variables, stochastic supergradient ascent updates the dual variable $\bblambda$ using a batch of $B$ samples
\begin{align*}
  d\bblambda_i =\frac{1}{B} \sum_{n=1}^B \ell_i (f_{\bbtheta}(x_n),y_n) - \bbzeta_i  \; \text{ for  } i = 1 \dots T_p.  
\end{align*}
Lastly, finding the parameters $\bbtheta$ that minimize the Lagrangian amounts to solving an empirical risk minimization problem with a time weighted loss, with weights given by the current multipliers $\bblambda$.

\section{Experiment Setup}~\label{app-exp-setup}
\textbf{Early Stopping} While the common practice in transformers for forecasting is to train with early stopping, we disable it for the constrained approach and train for a full 10 epochs, due to the slower convergence. Note that in the results presented in this work we keep early stopping for ERM.\\

\textbf{Tuning of dual learning rate and initialization parameters.} Preliminary exploration yielded consistently superior results with dual learning rate set to 0.01 and duals initialized to 1.0. All experiments reported in the paper were performed with this parameterization.

\textbf{Choice of $\bbepsilon$.} To choose an appropriate upper bound constraint for the stepwise losses, we perform a grid search of six values for every setting of dataset, model and prediction window, optimizing for validation MSE. The values for the search are the 25, 50, and 75th percentile of the training and validation errors of each model trained with ERM. The values reported in Tables~\ref{tab:mse_all_pt1}, ~\ref{tab:mse_all_pt2},~\ref{tab:window_std_pt1} and~\ref{tab:window_std_pt2} are for the optimal values of this grid search.

\textbf{Alternative choices of epsilon. } During preliminary experiments, we explored 




\subsection{Datasets}\label{app-datasets}


We use commmon multivariate, long-term forecasting benchmarks from the transformer time series literature. Here we provide a brief summary of each dataset. For more details about the data, refer to their respective sources.

\textbf{Weather}: Contains local climatological data for around 1,600 U.S. locations from 2010 to 2013. It includes 11 climate features, and the target is ``Wet Bulb Temp". The data is available at \url{https://www.ncei.noaa.gov/
data/local-climatological-data/}.

\textbf{ETT(Electricity Transformer Temperature)}: Records of transformer oil temperature and six power load features at 1-hour and 15-minute intervals across two years from two separated counties in China. The data was collected by~\cite{2021informer}.

\textbf{ECL (Electricity Consuming Load)}: Includes hourly electricity consumption of 321 clients over two years, with the target being the 'MT 320' consumption value. The data is available at \url{https://archive.ics.uci.edu/ml/
datasets/ElectricityLoadDiagrams20112014}.

\textbf{Exchange Rate}: Contains the daily exchange rates of Australia, British, Canada, Switzerland, China, Japan, New Zealand and Singapore from 1990 to 2016. The data was collected by~\cite{2018ExchangeLSTN}.

\textbf{ILI (National Illness)}: Contains weekly influenza-like illness (ILI) patients' data, showing the ratio of ILI patients to the total number of patients from 2002 to 2021. The data is available at \url{https://gis.cdc.gov/grasp/fluview/fluportaldashboard.html}.

\textbf{Traffic}: Contains hourly road occupancy rates measured by sensors in San Francisco Bay area freeways. The data is available at \url{https://pems.dot.ca.gov/}.

\section{Extended Experiment Results}\label{app-exp-results}

The unnormalized results used to compute the relative metrics presented in Section~\ref{sec:exps} are presented in Tables~\ref{tab:mse_all_pt1} and~\ref{tab:mse_all_pt2}, containing the MSE of ERM, constrained and resilient runs, and Tables~\ref{tab:window_std_pt1} and~\ref{tab:window_std_pt2} contain Window STDs. While the overall effect can be better appreciated across all settings in Figures~\ref{fig:histograms} and~\ref{fig:boxplots}, the detailed tables also showcase how the loss shaping effect is observed across a wide variety of dataset and model combinations.

\begin{figure}[h]
        \centering
         \includegraphics[width=0.9\linewidth]{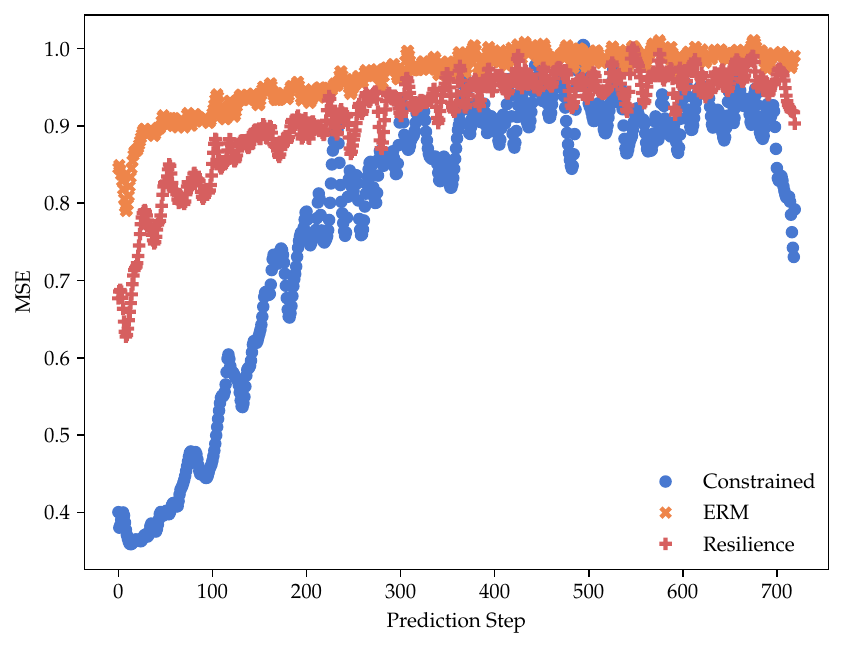}
    \caption{The anomaly of Figure~\ref{fig:histograms}. Test MSE of ERM, constrained and resilient runs. }\label{fig:anomaly}
\end{figure}

\begin{sidewaystable*}[htbp]
\footnotesize
\resizebox{0.9\textwidth}{!}{
\begin{tabular}{cc|ccc|ccc|ccc|ccc|ccc|ccc|ccc|ccc}
\toprule
 &  & \multicolumn{3}{c}{Autoformer} & \multicolumn{3}{c}{FiLM} & \multicolumn{3}{c}{Informer} & \multicolumn{3}{c}{Nonstationary Transformer} \\
 &  & ERM & Ours & Ours+R & ERM & Ours & Ours+R & ERM & Ours & Ours+R & ERM & Ours & Ours+R \\
\midrule
\multirow[t]{4}{*}{\multirow{4}{*}{\rotatebox{90}{Etth1}}} & 96 & 0.679 & 0.644 & \textbf{0.606} & 0.468 & \textbf{0.459} & 0.514 & 0.890 & \textbf{0.800} & 0.812 & 0.623 & 0.623 & \textbf{0.585} \\
 & 192 & 0.869 & \textbf{0.838} & 1.085 & 0.524 & \textbf{0.514} & 0.561 & 1.179 & 1.040 & \textbf{0.987} & 0.647 & 0.647 & \textbf{0.611} \\
 & 336 & 0.853 & 0.896 & \textbf{0.653} & 0.578 & \textbf{0.567} & 0.612 & 1.237 & \textbf{1.232} & 1.333 & 0.853 & \textbf{0.819} & 0.842 \\
 & 720 & 1.107 & \textbf{1.052} & 1.274 & 0.751 & \textbf{0.732} & 0.769 & 1.278 & \textbf{1.277} & 1.285 & 0.967 & 0.974 & \textbf{0.899} \\
\cline{1-14}
\multirow[t]{4}{*}{\multirow{4}{*}{\rotatebox{90}{Etth2}}} & 96 & 0.314 & 0.261 & \textbf{0.219} & 0.189 & \textbf{0.188} & 0.200 & 0.433 & \textbf{0.349} & 0.349 & 0.239 & \textbf{0.238} & 0.254 \\
 & 192 & 0.340 & \textbf{0.319} & 0.507 & 0.232 & \textbf{0.231} & 0.243 & 0.648 & 0.488 & \textbf{0.478} & 0.294 & \textbf{0.289} & 0.294 \\
 & 336 & 0.399 & \textbf{0.334} & 0.355 & 0.266 & \textbf{0.264} & 0.275 & 0.675 & 0.462 & \textbf{0.441} & \textbf{0.296} & 0.296 & 0.296 \\
 & 720 & 0.436 & 0.470 & \textbf{0.339} & 0.343 & \textbf{0.340} & 0.351 & 0.794 & 0.665 & \textbf{0.653} & 0.316 & \textbf{0.312} & 0.322 \\
\cline{1-14}
\multirow[t]{4}{*}{\multirow{4}{*}{\rotatebox{90}{Ettm1}}} & 96 & 0.605 & \textbf{0.588} & 0.601 & \textbf{0.363} & 0.369 & 0.374 & 0.646 & \textbf{0.571} & 0.585 & \textbf{0.535} & \textbf{0.535} & 0.634 \\
 & 192 & 0.828 & \textbf{0.641} & 0.760 & \textbf{0.431} & 0.436 & 0.442 & 0.768 & \textbf{0.577} & 0.585 & 0.592 & \textbf{0.581} & 0.649 \\
 & 336 & 0.685 & \textbf{0.658} & 0.749 & \textbf{0.502} & 0.505 & 0.512 & 0.857 & \textbf{0.842} & 0.848 & \textbf{0.661} & 0.666 & 0.694 \\
 & 720 & 0.746 & \textbf{0.726} & 0.734 & \textbf{0.577} & 0.579 & 0.590 & 0.929 & 0.922 & \textbf{0.900} & \textbf{0.748} & \textbf{0.748} & 0.818 \\
\cline{1-14}
\multirow[t]{4}{*}{\multirow{4}{*}{\rotatebox{90}{Ettm2}}} & 96 & 0.168 & 0.165 & \textbf{0.148} & \textbf{0.122} & 0.123 & 0.125 & 0.190 & 0.141 & \textbf{0.137} & 0.139 & \textbf{0.135} & 0.141 \\
 & 192 & 0.191 & 0.172 & \textbf{0.169} & \textbf{0.150} & 0.152 & 0.153 & 0.223 & \textbf{0.188} & 0.200 & 0.178 & \textbf{0.175} & 0.187 \\
 & 336 & 0.214 & \textbf{0.197} & 0.224 & \textbf{0.183} & 0.185 & 0.186 & 0.289 & \textbf{0.282} & 0.309 & 0.246 & \textbf{0.230} & 0.251 \\
 & 720 & 0.271 & \textbf{0.260} & 0.336 & \textbf{0.242} & 0.243 & 0.244 & 0.572 & \textbf{0.533} & 0.558 & 0.328 & 0.328 & \textbf{0.306} \\
\cline{1-14}
\multirow[t]{4}{*}{\multirow{4}{*}{\rotatebox{90}{ECL}}} & 96 & 0.204 & \textbf{0.200} & 0.220 & 0.239 & \textbf{0.198} & 0.217 & 0.325 & \textbf{0.319} & 0.342 & 0.178 & 0.182 & \textbf{0.175} \\
 & 192 & \textbf{0.215} & 0.226 & 0.258 & 0.266 & \textbf{0.200} & 0.221 & 0.352 & \textbf{0.341} & 0.360 & \textbf{0.186} & 0.191 & 0.190 \\
 & 336 & 0.334 & \textbf{0.249} & 0.287 & 0.306 & \textbf{0.219} & 0.239 & 0.354 & \textbf{0.348} & 0.371 & \textbf{0.204} & 0.204 & 0.206 \\
 & 720 & \textbf{0.252} & 0.270 & 0.410 & 0.351 & \textbf{0.281} & 0.294 & 0.385 & \textbf{0.377} & 0.438 & 0.233 & \textbf{0.229} & 0.230 \\
\cline{1-14}
\multirow[t]{4}{*}{\multirow{4}{*}{\rotatebox{90}{Exchange}}} & 96 & \textbf{0.155} & 0.179 & 0.163 & 0.162 & \textbf{0.094} & 0.100 & 0.880 & \textbf{0.872} & 0.910 & 0.160 & 0.152 & \textbf{0.139} \\
 & 192 & 2.052 & \textbf{0.266} & 0.267 & 0.256 & \textbf{0.182} & 0.187 & 1.113 & \textbf{1.110} & 1.131 & \textbf{0.268} & 0.290 & 0.273 \\
 & 336 & 3.442 & 0.446 & \textbf{0.442} & 0.542 & \textbf{0.319} & 0.323 & 1.633 & 1.088 & \textbf{1.080} & \textbf{0.412} & 0.573 & 0.488 \\
 & 720 & 1.852 & 1.526 & \textbf{0.995} & 1.153 & \textbf{0.820} & 0.822 & 2.961 & 1.170 & \textbf{1.166} & 1.342 & \textbf{1.120} & 1.123 \\
\cline{1-14}
\multirow[t]{4}{*}{\multirow{4}{*}{\rotatebox{90}{Illness}}} & 24 & \textbf{3.180} & 3.398 & 3.454 & 3.441 & \textbf{3.436} & 3.452 & 5.213 & \textbf{5.174} & 5.264 & 2.476 & \textbf{2.275} & 2.305 \\
 & 36 & 3.216 & 3.213 & \textbf{3.182} & 3.929 & 3.920 & \textbf{3.918} & 5.785 & 5.792 & \textbf{5.427} & \textbf{2.257} & 2.413 & 2.462 \\
 & 48 & 3.013 & \textbf{2.738} & 2.746 & 3.365 & \textbf{3.360} & 3.367 & 5.500 & \textbf{5.496} & 5.686 & 2.703 & \textbf{2.546} & 2.605 \\
 & 60 & 3.148 & 3.106 & \textbf{2.875} & 3.783 & \textbf{3.780} & 3.782 & 5.772 & 5.783 & \textbf{5.703} & 2.634 & \textbf{2.587} & 2.606 \\
\cline{1-14}
\multirow[t]{4}{*}{\multirow{4}{*}{\rotatebox{90}{Traffic}}} & 96 & 0.633 & \textbf{0.615} & 0.691 & \textbf{0.411} & 0.652 & 0.695 & 0.726 & \textbf{0.716} & 0.886 & 0.634 & \textbf{0.604} & 0.625 \\
 & 192 & 0.613 & \textbf{0.602} & 0.679 & \textbf{0.408} & 0.605 & 0.651 & \textbf{0.736} & 0.738 & 0.776 & \textbf{0.624} & 0.631 & 0.649 \\
 & 336 & 0.612 & \textbf{0.606} & 0.622 & \textbf{0.426} & 0.616 & 0.667 & 0.830 & \textbf{0.821} & 0.891 & \textbf{0.625} & 0.646 & 0.653 \\
 & 720 & \textbf{0.675} & 0.677 & 0.785 & \textbf{0.526} & 0.711 & 0.775 & 1.016 & \textbf{0.885} & 1.016 & 0.660 & \textbf{0.657} & 0.699 \\
\cline{1-14}
\multirow[t]{4}{*}{\multirow{4}{*}{\rotatebox{90}{Weather}}} & 96 & 0.375 & \textbf{0.300} & 0.314 & \textbf{0.196} & 0.198 & 0.199 & 0.435 & \textbf{0.395} & 0.441 & \textbf{0.203} & 0.204 & 0.206 \\
 & 192 & \textbf{0.303} & 0.311 & 0.322 & \textbf{0.234} & 0.240 & 0.241 & 0.515 & \textbf{0.447} & 0.491 & \textbf{0.271} & 0.283 & 0.286 \\
 & 336 & \textbf{0.359} & 0.374 & 0.402 & \textbf{0.278} & 0.290 & 0.291 & \textbf{0.600} & 0.656 & 0.600 & 0.383 & \textbf{0.356} & 0.375 \\
 & 720 & 0.443 & \textbf{0.442} & 0.447 & \textbf{0.342} & 0.361 & 0.362 & 1.100 & \textbf{1.051} & 1.198 & 0.454 & 0.466 & \textbf{0.438} \\
\cline{1-14}
\bottomrule
\end{tabular}

}
\caption{MSE of ERM, constrained (Ours) and resilient runs (Ours+R) for each model and dataset, part 1/2, for each prediction length. }
\label{tab:mse_all_pt1}
\end{sidewaystable*}

\newpage

\begin{sidewaystable*}[htbp]
\footnotesize
\resizebox{0.88\textwidth}{!}{
\begin{tabular}{cc|ccc|ccc|ccc|ccc|ccc|ccc|ccc|ccc}
\toprule
 &  & \multicolumn{3}{c}{Pyraformer} & \multicolumn{3}{c}{Reformer} & \multicolumn{3}{c}{Transformer} & \multicolumn{3}{c}{iTransformer} \\
 &  & ERM & Ours & Ours+R & ERM & Ours & Ours+R & ERM & Ours & Ours+R & ERM & Ours & Ours+R \\
\midrule
\multirow[t]{4}{*}{\multirow{4}{*}{\rotatebox{90}{Etth1}}} & 96 & \textbf{0.569} & \textbf{0.569} & 0.572 & 0.789 & \textbf{0.691} & 0.727 & 0.866 & 0.770 & \textbf{0.739} & 0.459 & 0.459 & \textbf{0.440} \\
 & 192 & \textbf{0.618} & \textbf{0.618} & 0.636 & \textbf{0.818} & 0.820 & 0.867 & 0.988 & \textbf{0.780} & 0.865 & 0.539 & 0.539 & \textbf{0.535} \\
 & 336 & 0.756 & 0.750 & \textbf{0.748} & \textbf{0.833} & 0.941 & 1.006 & \textbf{0.941} & 0.968 & 1.108 & \textbf{0.594} & 0.601 & 0.602 \\
 & 720 & 0.994 & 0.971 & \textbf{0.960} & \textbf{0.911} & 1.038 & 1.164 & 0.989 & \textbf{0.954} & 1.113 & \textbf{0.747} & \textbf{0.747} & 0.755 \\
\cline{1-14}
\multirow[t]{4}{*}{\multirow{4}{*}{\rotatebox{90}{Etth2}}} & 96 & 0.277 & 0.277 & \textbf{0.251} & 0.562 & 0.309 & \textbf{0.307} & 0.274 & \textbf{0.231} & 0.233 & 0.203 & 0.203 & \textbf{0.190} \\
 & 192 & 0.583 & 0.588 & \textbf{0.548} & 1.246 & \textbf{0.359} & 0.374 & \textbf{0.316} & 0.327 & 0.320 & 0.242 & 0.247 & \textbf{0.234} \\
 & 336 & 0.597 & 0.595 & \textbf{0.541} & 1.327 & \textbf{0.442} & 0.453 & \textbf{0.378} & 0.392 & 0.383 & 0.271 & 0.279 & \textbf{0.271} \\
 & 720 & 0.660 & 0.660 & \textbf{0.524} & 2.062 & 0.711 & \textbf{0.615} & \textbf{0.381} & 0.490 & 0.496 & 0.323 & 0.327 & \textbf{0.321} \\
\cline{1-14}
\multirow[t]{4}{*}{\multirow{4}{*}{\rotatebox{90}{Ettm1}}} & 96 & 0.542 & 0.542 & \textbf{0.495} & 0.697 & \textbf{0.517} & 0.518 & 0.569 & \textbf{0.565} & 0.579 & 0.414 & 0.414 & \textbf{0.383} \\
 & 192 & 0.607 & 0.607 & \textbf{0.522} & 0.830 & 0.618 & \textbf{0.613} & \textbf{0.637} & 0.650 & 0.689 & 0.476 & 0.476 & \textbf{0.461} \\
 & 336 & 0.658 & 0.658 & \textbf{0.595} & 0.860 & 0.707 & \textbf{0.707} & 0.735 & \textbf{0.713} & 0.773 & \textbf{0.535} & \textbf{0.535} & 0.539 \\
 & 720 & 0.760 & \textbf{0.744} & 0.757 & 0.892 & 0.868 & \textbf{0.861} & 0.989 & 0.865 & \textbf{0.823} & 0.608 & 0.608 & \textbf{0.602} \\
\cline{1-14}
\multirow[t]{4}{*}{\multirow{4}{*}{\rotatebox{90}{Ettm2}}} & 96 & 0.143 & 0.143 & \textbf{0.135} & 0.183 & 0.173 & \textbf{0.172} & 0.162 & 0.135 & \textbf{0.133} & 0.134 & 0.134 & \textbf{0.126} \\
 & 192 & 0.212 & 0.212 & \textbf{0.188} & 0.297 & 0.241 & \textbf{0.239} & 0.213 & \textbf{0.203} & 0.209 & 0.178 & 0.178 & \textbf{0.161} \\
 & 336 & \textbf{0.335} & 0.371 & 0.351 & 0.739 & 0.319 & \textbf{0.314} & 0.281 & 0.293 & \textbf{0.259} & 0.212 & 0.211 & \textbf{0.210} \\
 & 720 & 0.508 & 0.508 & \textbf{0.422} & 1.878 & \textbf{0.549} & 0.566 & 0.413 & 0.443 & \textbf{0.398} & 0.271 & \textbf{0.265} & 0.267 \\
\cline{1-14}
\multirow[t]{4}{*}{\multirow{4}{*}{\rotatebox{90}{ECL}}} & 96 & 0.287 & \textbf{0.280} & 0.303 & 0.299 & \textbf{0.297} & 0.303 & 0.262 & \textbf{0.259} & 0.278 & \textbf{0.148} & 0.159 & 0.171 \\
 & 192 & 0.294 & \textbf{0.291} & 0.320 & 0.332 & \textbf{0.326} & 0.332 & 0.281 & \textbf{0.274} & 0.321 & \textbf{0.164} & 0.173 & 0.183 \\
 & 336 & 0.313 & \textbf{0.301} & 0.324 & 0.359 & 0.365 & \textbf{0.329} & \textbf{0.280} & 0.287 & 0.297 & \textbf{0.178} & 0.190 & 0.200 \\
 & 720 & 0.310 & \textbf{0.298} & 0.324 & \textbf{0.316} & 0.316 & 0.326 & \textbf{0.289} & 0.293 & 0.312 & \textbf{0.208} & 0.227 & 0.240 \\
\cline{1-14}
\multirow[t]{4}{*}{\multirow{4}{*}{\rotatebox{90}{Exchange}}} & 96 & 0.802 & 0.774 & \textbf{0.584} & \textbf{0.960} & 1.025 & 0.981 & 0.691 & 0.593 & \textbf{0.556} & \textbf{0.087} & 0.117 & 0.099 \\
 & 192 & 1.216 & 1.260 & \textbf{0.988} & \textbf{1.398} & 1.402 & 1.685 & 1.235 & 1.253 & \textbf{1.174} & \textbf{0.180} & 0.226 & 0.199 \\
 & 336 & 1.666 & 1.388 & \textbf{1.319} & 1.909 & \textbf{1.632} & 1.924 & 1.421 & \textbf{1.362} & 1.649 & \textbf{0.333} & 0.364 & 0.344 \\
 & 720 & 2.051 & 1.725 & \textbf{1.686} & 2.095 & \textbf{2.063} & 2.114 & 2.197 & 1.029 & \textbf{1.015} & \textbf{0.865} & 0.986 & 0.920 \\
\cline{1-14}
\multirow[t]{4}{*}{\multirow{4}{*}{\rotatebox{90}{Illness}}} & 24 & 4.724 & 4.724 & \textbf{4.721} & 3.948 & 3.944 & \textbf{3.901} & \textbf{5.203} & 5.270 & 5.959 & 1.935 & \textbf{1.918} & 1.942 \\
 & 36 & 4.972 & \textbf{4.924} & 4.965 & 4.069 & \textbf{4.069} & 4.218 & 4.571 & \textbf{4.528} & 6.038 & 1.865 & \textbf{1.862} & 1.871 \\
 & 48 & 4.686 & 4.722 & \textbf{4.677} & 4.432 & 4.433 & \textbf{4.335} & 5.166 & \textbf{5.073} & 5.678 & \textbf{1.942} & 1.945 & 1.949 \\
 & 60 & 4.892 & \textbf{4.889} & 4.893 & 4.574 & \textbf{4.564} & 4.575 & 5.572 & \textbf{5.482} & 5.673 & 1.849 & \textbf{1.849} & 1.860 \\
\cline{1-14}
\multirow[t]{4}{*}{\multirow{4}{*}{\rotatebox{90}{Traffic}}} & 96 & 0.691 & \textbf{0.678} & 0.704 & 0.698 & \textbf{0.693} & 0.702 & \textbf{0.639} & 0.640 & 0.649 & \textbf{0.392} & 0.435 & 0.459 \\
 & 192 & 0.676 & \textbf{0.660} & 0.694 & 0.690 & \textbf{0.684} & 0.703 & 0.642 & \textbf{0.635} & 0.651 & \textbf{0.411} & 0.452 & 0.475 \\
 & 336 & 0.694 & \textbf{0.671} & 0.713 & 0.690 & \textbf{0.686} & 0.704 & 0.666 & \textbf{0.663} & 0.669 & \textbf{0.424} & 0.471 & 0.494 \\
 & 720 & 0.713 & \textbf{0.708} & 0.743 & 0.682 & \textbf{0.678} & 0.704 & 0.698 & \textbf{0.693} & 0.696 & \textbf{0.454} & 0.508 & 0.536 \\
\cline{1-14}
\multirow[t]{4}{*}{\multirow{4}{*}{\rotatebox{90}{Weather}}} & 96 & 0.217 & 0.218 & \textbf{0.213} & 0.422 & 0.334 & \textbf{0.331} & 0.448 & 0.462 & \textbf{0.435} & \textbf{0.174} & 0.176 & 0.177 \\
 & 192 & 0.287 & \textbf{0.269} & 0.270 & 0.703 & \textbf{0.425} & 0.483 & 0.667 & \textbf{0.593} & 0.640 & 0.226 & \textbf{0.224} & 0.225 \\
 & 336 & 0.341 & \textbf{0.325} & 0.330 & 1.010 & \textbf{0.588} & 0.995 & \textbf{0.749} & 0.901 & 0.915 & 0.285 & 0.284 & \textbf{0.282} \\
 & 720 & 0.460 & \textbf{0.417} & 0.423 & 0.963 & \textbf{0.788} & 0.914 & 1.002 & \textbf{0.924} & 1.072 & 0.363 & 0.363 & \textbf{0.362} \\
\cline{1-14}
\bottomrule
\end{tabular}

}
\caption{MSE of ERM, constrained (Ours) and resilient runs (Ours+R) for each model and dataset, part 2/2, for each prediction length. }
\label{tab:mse_all_pt2}
\end{sidewaystable*}

\begin{sidewaystable*}[htbp]
\footnotesize
\resizebox{0.9\textwidth}{!}{

\begin{tabular}{cc|ccc|ccc|ccc|ccc|ccc|ccc|ccc|ccc}
\toprule
 &  & \multicolumn{3}{c}{Autoformer} & \multicolumn{3}{c}{FiLM} & \multicolumn{3}{c}{Informer} & \multicolumn{3}{c}{Nonstationary Transformer} \\
 &  & ERM & Ours & Ours+R & ERM & Ours & Ours+R & ERM & Ours & Ours+R & ERM & Ours & Ours+R \\
\midrule
\multirow[t]{4}{*}{\multirow{4}{*}{\rotatebox{90}{Etth1}}} & 96 & 0.024 & 0.032 & \textbf{0.019} & 0.070 & 0.064 & \textbf{0.052} & 0.183 & 0.122 & \textbf{0.114} & 0.060 & 0.060 & \textbf{0.053} \\
 & 192 & \textbf{0.139} & 0.141 & 0.338 & 0.073 & 0.064 & \textbf{0.052} & 0.211 & 0.183 & \textbf{0.128} & 0.086 & 0.086 & \textbf{0.081} \\
 & 336 & 0.083 & 0.109 & \textbf{0.028} & 0.075 & 0.063 & \textbf{0.053} & 0.181 & \textbf{0.156} & 0.200 & \textbf{0.187} & 0.207 & 0.209 \\
 & 720 & 0.557 & \textbf{0.278} & 1.208 & 0.142 & 0.134 & \textbf{0.126} & 0.170 & \textbf{0.169} & 0.192 & 0.171 & \textbf{0.160} & 0.164 \\
\cline{1-14}
\multirow[t]{4}{*}{\multirow{4}{*}{\rotatebox{90}{Etth2}}} & 96 & 0.041 & 0.029 & \textbf{0.028} & 0.046 & 0.043 & \textbf{0.041} & 0.138 & 0.110 & \textbf{0.109} & 0.070 & \textbf{0.069} & 0.072 \\
 & 192 & \textbf{0.015} & 0.023 & 0.047 & 0.053 & 0.050 & \textbf{0.048} & 0.251 & 0.155 & \textbf{0.147} & 0.059 & 0.057 & \textbf{0.055} \\
 & 336 & 0.079 & \textbf{0.020} & 0.073 & 0.054 & 0.051 & \textbf{0.049} & 0.203 & 0.097 & \textbf{0.088} & 0.045 & \textbf{0.042} & 0.045 \\
 & 720 & 0.071 & 0.088 & \textbf{0.045} & 0.071 & 0.067 & \textbf{0.066} & 0.248 & 0.142 & \textbf{0.118} & 0.047 & 0.036 & \textbf{0.035} \\
\cline{1-14}
\multirow[t]{4}{*}{\multirow{4}{*}{\rotatebox{90}{Ettm1}}} & 96 & 0.063 & \textbf{0.061} & 0.065 & 0.070 & 0.054 & \textbf{0.051} & 0.157 & \textbf{0.119} & 0.133 & \textbf{0.120} & \textbf{0.120} & 0.175 \\
 & 192 & 0.091 & 0.093 & \textbf{0.038} & 0.088 & 0.077 & \textbf{0.069} & 0.120 & \textbf{0.066} & 0.072 & 0.109 & \textbf{0.104} & 0.117 \\
 & 336 & 0.062 & 0.063 & \textbf{0.033} & 0.105 & 0.098 & \textbf{0.089} & 0.170 & 0.172 & \textbf{0.137} & \textbf{0.102} & 0.104 & 0.108 \\
 & 720 & 0.062 & 0.061 & \textbf{0.055} & 0.102 & 0.094 & \textbf{0.076} & 0.128 & 0.127 & \textbf{0.101} & \textbf{0.112} & \textbf{0.112} & 0.143 \\
\cline{1-14}
\multirow[t]{4}{*}{\multirow{4}{*}{\rotatebox{90}{Ettm2}}} & 96 & 0.034 & 0.024 & \textbf{0.020} & 0.028 & 0.025 & \textbf{0.022} & 0.064 & 0.030 & \textbf{0.029} & 0.036 & \textbf{0.032} & 0.035 \\
 & 192 & 0.027 & \textbf{0.025} & 0.026 & 0.035 & 0.032 & \textbf{0.030} & 0.065 & \textbf{0.050} & 0.051 & 0.051 & \textbf{0.049} & 0.054 \\
 & 336 & 0.041 & 0.038 & \textbf{0.032} & 0.048 & 0.046 & \textbf{0.044} & 0.080 & \textbf{0.076} & 0.080 & 0.082 & \textbf{0.070} & 0.080 \\
 & 720 & \textbf{0.052} & 0.061 & 0.072 & 0.064 & 0.061 & \textbf{0.059} & 0.258 & \textbf{0.222} & 0.228 & 0.087 & 0.087 & \textbf{0.084} \\
\cline{1-14}
\multirow[t]{4}{*}{\multirow{4}{*}{\rotatebox{90}{ECL}}} & 96 & 0.015 & 0.015 & \textbf{0.013} & 0.050 & 0.031 & \textbf{0.029} & 0.014 & \textbf{0.013} & 0.018 & 0.017 & \textbf{0.014} & 0.015 \\
 & 192 & 0.022 & 0.021 & \textbf{0.020} & 0.045 & 0.029 & \textbf{0.026} & 0.041 & 0.037 & \textbf{0.029} & 0.026 & \textbf{0.019} & 0.020 \\
 & 336 & 0.043 & 0.039 & \textbf{0.020} & 0.055 & 0.037 & \textbf{0.032} & 0.022 & \textbf{0.020} & 0.021 & 0.031 & \textbf{0.024} & 0.029 \\
 & 720 & 0.035 & \textbf{0.030} & 0.156 & 0.052 & 0.053 & \textbf{0.048} & 0.028 & \textbf{0.023} & 0.040 & 0.040 & 0.037 & \textbf{0.033} \\
\cline{1-14}
\multirow[t]{4}{*}{\multirow{4}{*}{\rotatebox{90}{Exchange}}} & 96 & 0.081 & 0.085 & \textbf{0.049} & 0.058 & 0.044 & \textbf{0.043} & 0.279 & \textbf{0.274} & 0.305 & 0.095 & 0.083 & \textbf{0.072} \\
 & 192 & 1.349 & \textbf{0.115} & 0.115 & 0.116 & 0.096 & \textbf{0.095} & 0.308 & \textbf{0.299} & 0.331 & \textbf{0.143} & 0.153 & 0.148 \\
 & 336 & 2.134 & 0.210 & \textbf{0.208} & 0.229 & 0.177 & \textbf{0.176} & 0.388 & 0.191 & \textbf{0.178} & \textbf{0.215} & 0.287 & 0.244 \\
 & 720 & 1.038 & 0.743 & \textbf{0.320} & 0.665 & \textbf{0.485} & 0.487 & 0.946 & 0.128 & \textbf{0.125} & 0.938 & \textbf{0.643} & 0.646 \\
\cline{1-14}
\multirow[t]{4}{*}{\multirow{4}{*}{\rotatebox{90}{Illness}}} & 24 & \textbf{0.113} & 0.140 & 0.203 & 0.607 & 0.602 & \textbf{0.599} & \textbf{0.438} & 0.472 & 0.510 & \textbf{0.361} & 0.517 & 0.515 \\
 & 36 & \textbf{0.197} & 0.213 & 0.342 & 0.557 & \textbf{0.547} & 0.549 & 0.700 & 0.699 & \textbf{0.658} & \textbf{0.279} & 0.500 & 0.513 \\
 & 48 & 0.308 & \textbf{0.237} & 0.376 & 0.618 & 0.615 & \textbf{0.614} & \textbf{0.651} & 0.725 & 0.932 & \textbf{0.243} & 0.474 & 0.457 \\
 & 60 & 0.561 & 0.566 & \textbf{0.398} & 0.405 & 0.403 & \textbf{0.403} & \textbf{1.029} & 1.037 & 1.077 & \textbf{0.485} & 0.514 & 0.500 \\
\cline{1-14}
\multirow[t]{4}{*}{\multirow{4}{*}{\rotatebox{90}{Traffic}}} & 96 & 0.023 & \textbf{0.019} & 0.024 & \textbf{0.026} & 0.121 & 0.117 & 0.030 & \textbf{0.025} & 0.057 & 0.018 & 0.014 & \textbf{0.011} \\
 & 192 & 0.036 & \textbf{0.026} & 0.047 & \textbf{0.025} & 0.125 & 0.117 & \textbf{0.038} & 0.048 & 0.042 & 0.021 & 0.017 & \textbf{0.017} \\
 & 336 & 0.031 & 0.029 & \textbf{0.027} & \textbf{0.031} & 0.133 & 0.125 & 0.077 & \textbf{0.068} & 0.085 & 0.021 & 0.026 & \textbf{0.018} \\
 & 720 & \textbf{0.036} & 0.040 & 0.064 & \textbf{0.054} & 0.133 & 0.118 & 0.071 & 0.066 & \textbf{0.062} & 0.029 & 0.032 & \textbf{0.023} \\
\cline{1-14}
\multirow[t]{4}{*}{\multirow{4}{*}{\rotatebox{90}{Weather}}} & 96 & 0.060 & 0.033 & \textbf{0.031} & 0.048 & \textbf{0.043} & 0.043 & 0.118 & 0.101 & \textbf{0.099} & 0.055 & \textbf{0.047} & 0.054 \\
 & 192 & 0.053 & 0.051 & \textbf{0.045} & 0.058 & \textbf{0.055} & 0.055 & 0.158 & \textbf{0.129} & 0.160 & 0.085 & \textbf{0.081} & 0.084 \\
 & 336 & 0.068 & \textbf{0.063} & 0.065 & \textbf{0.070} & 0.070 & 0.070 & 0.136 & \textbf{0.119} & 0.177 & 0.113 & \textbf{0.104} & 0.114 \\
 & 720 & 0.079 & 0.074 & \textbf{0.061} & \textbf{0.083} & 0.089 & 0.088 & 0.364 & \textbf{0.251} & 0.528 & 0.117 & \textbf{0.107} & 0.126 \\
\cline{1-14}
\bottomrule
\end{tabular}

}
\caption{Window STD of ERM, constrained (Ours) and resilient runs (Ours+R) for each model and dataset, part 1/2, for each prediction length. }
\label{tab:window_std_pt1}
\end{sidewaystable*}

\begin{sidewaystable*}[htbp]
\footnotesize
\resizebox{0.9\textwidth}{!}{

\begin{tabular}{cc|ccc|ccc|ccc|ccc|ccc|ccc|ccc|ccc}
\toprule
 &  & \multicolumn{3}{c}{Pyraformer} & \multicolumn{3}{c}{Reformer} & \multicolumn{3}{c}{Transformer} & \multicolumn{3}{c}{iTransformer} \\
 &  & 1ERM & 2Ours & 3Ours+R & 1ERM & 2Ours & 3Ours+R & 1ERM & 2Ours & 3Ours+R & 1ERM & 2Ours & 3Ours+R \\
\midrule
\multirow[t]{4}{*}{\multirow{4}{*}{\rotatebox{90}{Etth1}}} & 96 & 0.074 & 0.074 & \textbf{0.063} & 0.058 & \textbf{0.045} & 0.046 & 0.146 & 0.120 & \textbf{0.096} & 0.081 & 0.081 & \textbf{0.065} \\
 & 192 & 0.082 & 0.082 & \textbf{0.072} & 0.077 & \textbf{0.040} & 0.044 & 0.166 & 0.130 & \textbf{0.087} & 0.095 & 0.095 & \textbf{0.082} \\
 & 336 & 0.153 & \textbf{0.130} & 0.146 & 0.062 & \textbf{0.029} & 0.044 & 0.123 & 0.137 & \textbf{0.117} & 0.104 & \textbf{0.096} & 0.097 \\
 & 720 & 0.170 & 0.173 & \textbf{0.140} & 0.045 & \textbf{0.044} & 0.062 & \textbf{0.093} & 0.116 & 0.121 & 0.167 & 0.167 & \textbf{0.163} \\
\cline{1-14}
\multirow[t]{4}{*}{\multirow{4}{*}{\rotatebox{90}{Etth2}}} & 96 & 0.082 & 0.082 & \textbf{0.045} & 0.262 & 0.043 & \textbf{0.040} & 0.050 & 0.036 & \textbf{0.036} & 0.055 & 0.055 & \textbf{0.048} \\
 & 192 & 0.292 & 0.291 & \textbf{0.275} & 0.718 & \textbf{0.041} & 0.042 & \textbf{0.053} & 0.058 & 0.062 & 0.057 & 0.058 & \textbf{0.054} \\
 & 336 & 0.212 & 0.216 & \textbf{0.163} & 0.332 & \textbf{0.046} & 0.049 & 0.064 & 0.064 & \textbf{0.054} & 0.051 & 0.054 & \textbf{0.051} \\
 & 720 & 0.144 & 0.144 & \textbf{0.040} & 0.194 & 0.095 & \textbf{0.053} & 0.056 & \textbf{0.039} & 0.060 & 0.061 & 0.058 & \textbf{0.057} \\
\cline{1-14}
\multirow[t]{4}{*}{\multirow{4}{*}{\rotatebox{90}{Ettm1}}} & 96 & 0.127 & 0.127 & \textbf{0.108} & 0.137 & \textbf{0.082} & 0.086 & 0.114 & \textbf{0.108} & 0.111 & 0.105 & 0.105 & \textbf{0.093} \\
 & 192 & 0.104 & 0.104 & \textbf{0.082} & 0.116 & \textbf{0.059} & 0.062 & \textbf{0.098} & 0.101 & 0.124 & 0.104 & 0.104 & \textbf{0.099} \\
 & 336 & 0.117 & 0.117 & \textbf{0.087} & 0.105 & \textbf{0.069} & 0.072 & 0.101 & \textbf{0.099} & 0.104 & \textbf{0.113} & \textbf{0.113} & 0.117 \\
 & 720 & 0.122 & 0.113 & \textbf{0.106} & 0.074 & \textbf{0.044} & 0.050 & 0.152 & 0.124 & \textbf{0.107} & 0.106 & 0.106 & \textbf{0.102} \\
\cline{1-14}
\multirow[t]{4}{*}{\multirow{4}{*}{\rotatebox{90}{Ettm2}}} & 96 & 0.038 & 0.038 & \textbf{0.030} & 0.054 & 0.036 & \textbf{0.036} & 0.039 & 0.031 & \textbf{0.030} & 0.037 & 0.037 & \textbf{0.033} \\
 & 192 & 0.060 & 0.060 & \textbf{0.046} & 0.089 & \textbf{0.052} & 0.055 & 0.054 & \textbf{0.049} & 0.054 & 0.052 & 0.052 & \textbf{0.042} \\
 & 336 & \textbf{0.106} & 0.134 & 0.109 & 0.226 & \textbf{0.057} & 0.061 & 0.069 & 0.067 & \textbf{0.064} & 0.065 & 0.065 & \textbf{0.064} \\
 & 720 & 0.222 & 0.222 & \textbf{0.157} & 0.277 & 0.148 & \textbf{0.140} & 0.103 & 0.116 & \textbf{0.099} & \textbf{0.073} & 0.074 & 0.073 \\
\cline{1-14}
\multirow[t]{4}{*}{\multirow{4}{*}{\rotatebox{90}{ECL}}} & 96 & 0.019 & 0.015 & \textbf{0.015} & 0.011 & 0.011 & \textbf{0.008} & \textbf{0.007} & 0.008 & 0.008 & 0.024 & 0.023 & \textbf{0.023} \\
 & 192 & 0.022 & 0.019 & \textbf{0.016} & 0.020 & 0.020 & \textbf{0.012} & 0.018 & 0.015 & \textbf{0.013} & 0.028 & 0.025 & \textbf{0.025} \\
 & 336 & 0.034 & 0.025 & \textbf{0.021} & 0.022 & 0.023 & \textbf{0.011} & 0.020 & 0.022 & \textbf{0.018} & 0.034 & 0.034 & \textbf{0.033} \\
 & 720 & 0.015 & 0.016 & \textbf{0.015} & 0.007 & 0.007 & \textbf{0.007} & \textbf{0.013} & 0.018 & 0.018 & \textbf{0.041} & 0.045 & 0.043 \\
\cline{1-14}
\multirow[t]{4}{*}{\multirow{4}{*}{\rotatebox{90}{Exchange}}} & 96 & 0.175 & \textbf{0.105} & 0.130 & 0.116 & 0.121 & \textbf{0.098} & 0.168 & 0.105 & \textbf{0.088} & \textbf{0.049} & 0.072 & 0.059 \\
 & 192 & 0.360 & \textbf{0.139} & 0.278 & \textbf{0.132} & 0.148 & 0.196 & 0.377 & 0.385 & \textbf{0.373} & \textbf{0.103} & 0.137 & 0.115 \\
 & 336 & 0.179 & \textbf{0.095} & 0.203 & 0.193 & \textbf{0.154} & 0.240 & 0.218 & \textbf{0.211} & 0.304 & \textbf{0.194} & 0.223 & 0.204 \\
 & 720 & 0.788 & \textbf{0.163} & 0.170 & 0.314 & \textbf{0.280} & 0.344 & 0.611 & 0.169 & \textbf{0.165} & \textbf{0.527} & 0.605 & 0.561 \\
\cline{1-14}
\multirow[t]{4}{*}{\multirow{4}{*}{\rotatebox{90}{Illness}}} & 24 & 1.092 & 1.092 & \textbf{1.074} & 0.459 & 0.460 & \textbf{0.412} & \textbf{0.490} & 0.508 & 0.679 & 0.807 & \textbf{0.799} & 0.806 \\
 & 36 & 0.931 & \textbf{0.893} & 0.929 & 0.778 & 0.772 & \textbf{0.445} & 0.436 & \textbf{0.436} & 0.578 & 0.632 & 0.629 & \textbf{0.625} \\
 & 48 & 0.862 & 0.882 & \textbf{0.855} & 0.425 & 0.423 & \textbf{0.407} & 0.403 & \textbf{0.395} & 0.680 & \textbf{0.596} & 0.602 & 0.599 \\
 & 60 & 0.781 & 0.783 & \textbf{0.772} & 0.397 & 0.398 & \textbf{0.364} & 0.562 & 0.535 & \textbf{0.457} & 0.579 & 0.579 & \textbf{0.573} \\
\cline{1-14}
\multirow[t]{4}{*}{\multirow{4}{*}{\rotatebox{90}{Traffic}}} & 96 & 0.032 & 0.026 & \textbf{0.021} & 0.021 & 0.021 & \textbf{0.020} & 0.011 & \textbf{0.010} & 0.010 & \textbf{0.032} & 0.034 & 0.032 \\
 & 192 & 0.016 & \textbf{0.013} & 0.014 & 0.022 & 0.020 & \textbf{0.019} & 0.019 & 0.015 & \textbf{0.015} & \textbf{0.029} & 0.031 & 0.029 \\
 & 336 & 0.015 & \textbf{0.011} & 0.013 & 0.017 & 0.016 & \textbf{0.014} & 0.025 & 0.021 & \textbf{0.019} & \textbf{0.030} & 0.035 & 0.033 \\
 & 720 & 0.016 & 0.016 & \textbf{0.016} & 0.012 & \textbf{0.010} & 0.011 & 0.055 & 0.057 & \textbf{0.038} & \textbf{0.038} & 0.042 & 0.041 \\
\cline{1-14}
\multirow[t]{4}{*}{\multirow{4}{*}{\rotatebox{90}{Weather}}} & 96 & 0.056 & 0.053 & \textbf{0.052} & 0.102 & \textbf{0.086} & 0.088 & 0.127 & 0.126 & \textbf{0.121} & 0.047 & \textbf{0.045} & 0.046 \\
 & 192 & 0.096 & 0.085 & \textbf{0.084} & 0.142 & \textbf{0.117} & 0.157 & 0.243 & \textbf{0.206} & 0.242 & 0.062 & 0.061 & \textbf{0.061} \\
 & 336 & 0.107 & \textbf{0.098} & 0.104 & \textbf{0.141} & 0.158 & 0.191 & \textbf{0.225} & 0.295 & 0.294 & 0.081 & 0.079 & \textbf{0.079} \\
 & 720 & 0.156 & \textbf{0.132} & 0.134 & \textbf{0.039} & 0.184 & 0.067 & 0.289 & \textbf{0.276} & 0.321 & 0.097 & 0.094 & \textbf{0.093} \\
\cline{1-14}
\bottomrule
\end{tabular}

}
\caption{Window STD of ERM, constrained (Ours) and resilient runs (Ours+R) for each model and dataset, part 2/2, for each prediction length. }
\label{tab:window_std_pt2}
\end{sidewaystable*}

\begin{figure*}[t]
    \centering
    \includegraphics[width=0.95\linewidth]{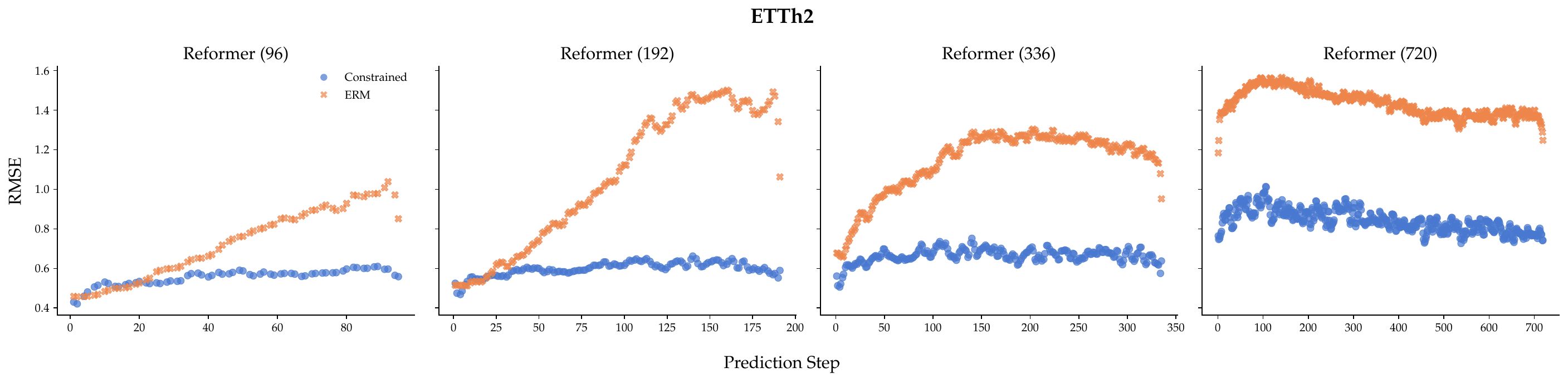}
    \includegraphics[width=0.95\linewidth]{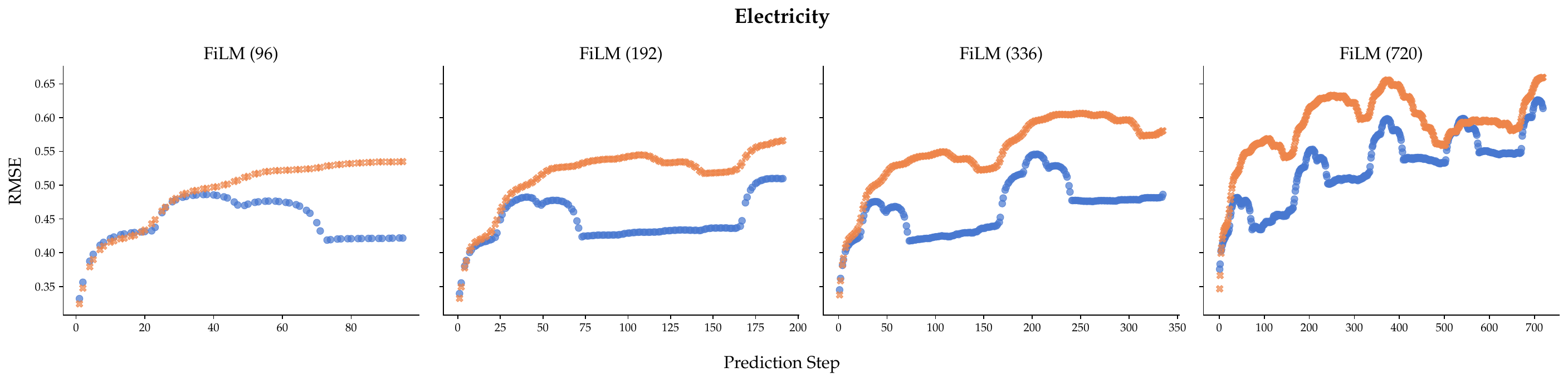}
    \includegraphics[width=0.95\linewidth]{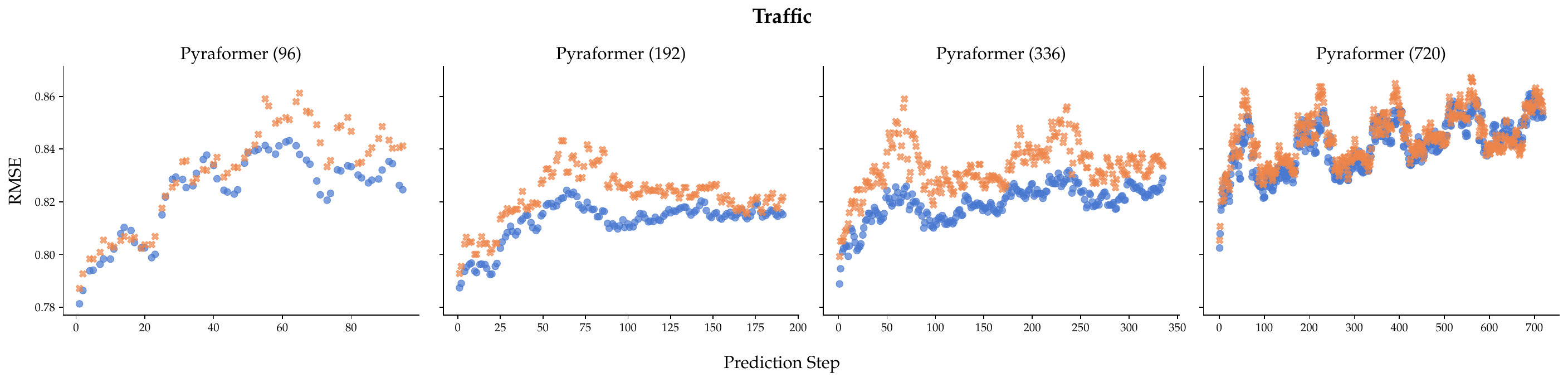}
    \centering
    \caption{Test MSE for ERM and constrained runs for three different settings: (a) Model: Reformer, Dataset: ETTh2, (b) Model: FiLM, Dataset: Traffic, and (c) Model: Pyraformer, Dataset: Traffic. Each plot column shows a different predictive window length, each row shows a different dataset-model selection.}
    \label{fig:full_plots}
\end{figure*}

\begin{figure*}[t]
    \centering
    \includegraphics[width=0.95\linewidth]{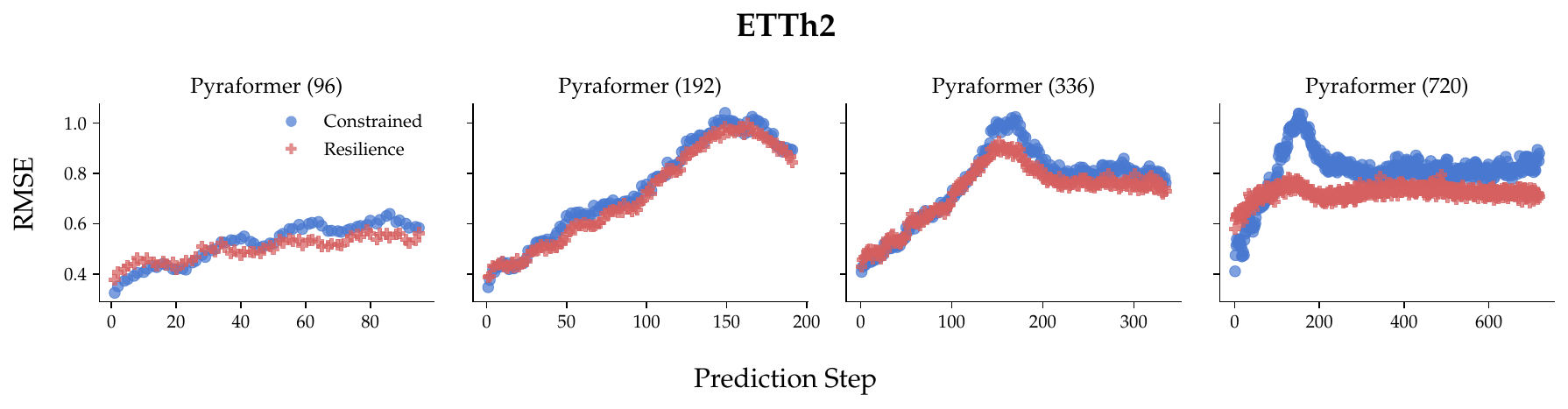}
    \includegraphics[width=0.95\linewidth]{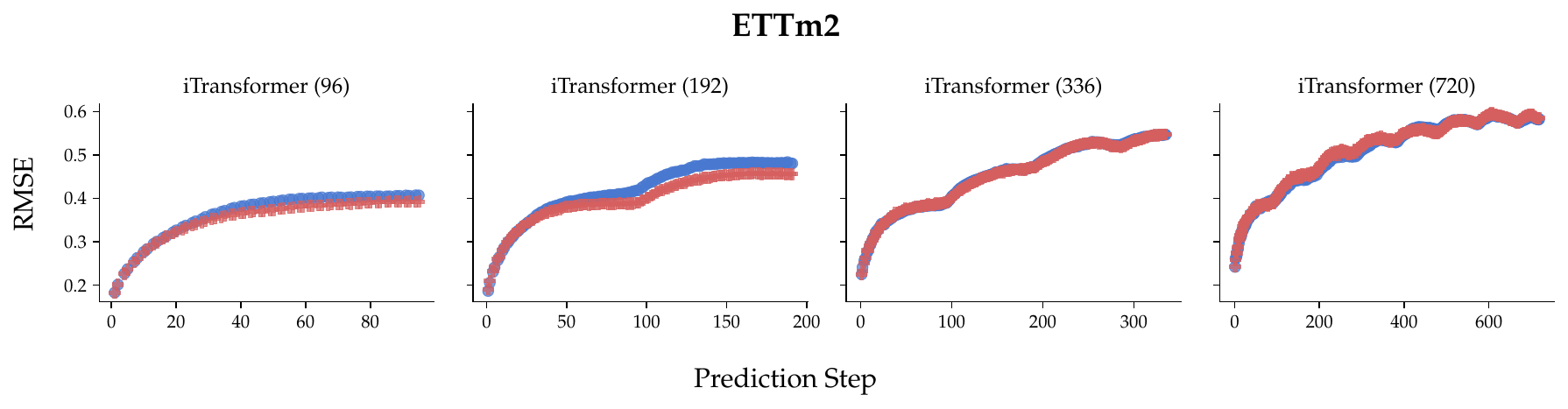}
    \includegraphics[width=0.95\linewidth]{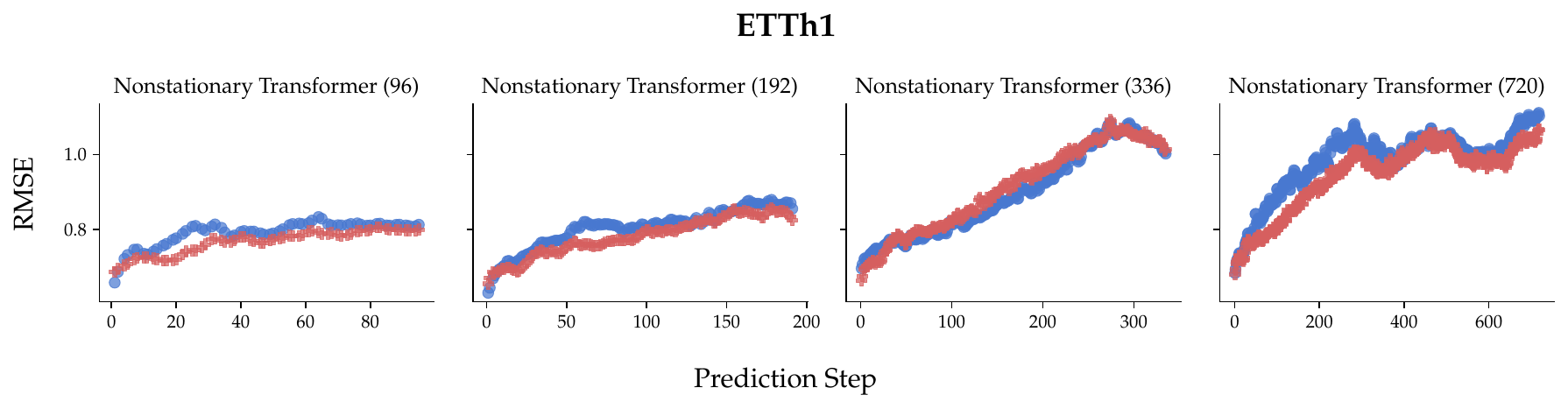}
    \centering
    \caption{Test MSE for constrained and resilient runs for three different settings: (a) Model: Pyraformer, Dataset: ETTh2, (b) Model: iTransformer, Dataset: ETTm2, and (c) Model: Nonstationary Transformer, Dataset: Exchange Rate. Each plot column shows a different predictive window length, each row shows a different dataset-model selection.}
    \label{fig:pred_len_plots_resilience}
\end{figure*}

\subsection{Loss Shape Visualization}\label{app-loss-shapes}
In this section, we take a closer look at some examples of loss distributions. As was shown in Section~\ref{subsec:loss-reshaping}, the loss shaping effect is consistent as the predictive window increases. This is illustrated in Figure~\ref{fig:full_plots}, where we plot the loss landscapes of three architectures tested on different datasets with increasing predictive window lengths. In all settings, a significant loss shaping effect is observed throughout the window, with the exception of the longest Pyraformer window (720), where the change is less pronounced. 

The particular loss shape remains idiosyncratic to each dataset, model, and predictive window length combination. In settings (a) and (b) of Figure~\ref{fig:full_plots}, a similar distribution pattern can be observed as the predictive window grows longer, while, in setting (c), the shape generated by the constrained model slightly varies with the window length.

\begin{figure}[!ht]
        \centering
         \includegraphics[width=0.95\linewidth]{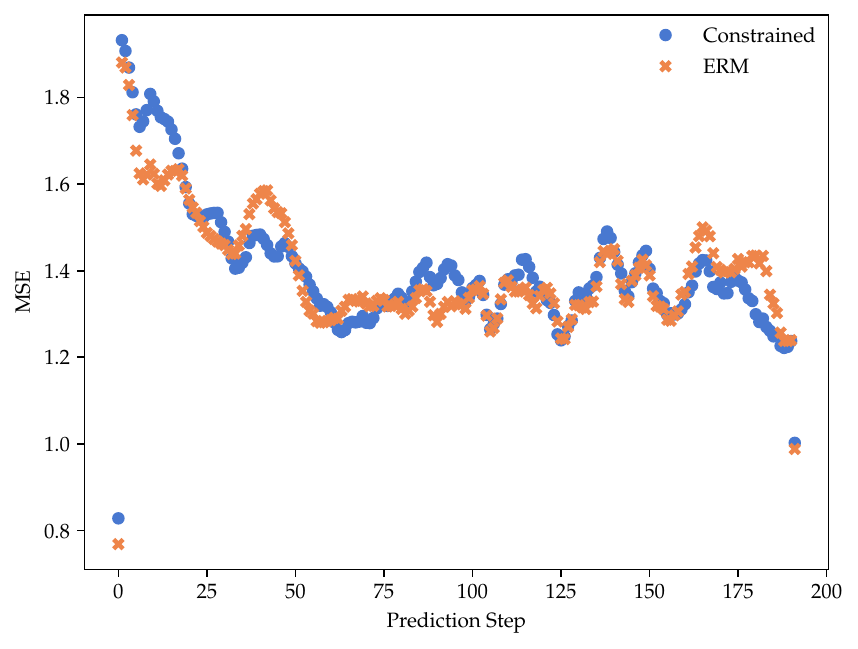}
    \caption{Test MSE of ERM and constrained models for Reformer on Exchange Rate, with window length of 192}\label{fig:downwards}
\end{figure}

Finally, we also observe that, more often than not, loss landscapes are not linear, let alone constant. Moreover, in rare cases, such as Reformer on Exchange Rate with prediction window of 192 steps (Figure~\ref{fig:downwards}), the stepwise loss surprisingly trends \textit{downwards}. This motivates further experimentation on different kinds of constraints that are better suited for a particular learning problem. 

\subsection{Resilient Loss Shape Plots}

We have presented extensive evidence that the resilient loss shaping effect is also consistent across different settings and predictive window lengths, and often times finds better shapes than non-resilient constrained learning. In Figure~\ref{fig:pred_len_plots_resilience},  we present three more loss distributions to illustrate this. In all window lengths of the three cases, except for Nonstationary Transformer (336), resilient learning results in flatter loss shapes or lower MSEs. Case (a) is worth noting: by finding the optimal relaxation of constraints, resilient is able to trade off a higher error in the first timesteps for a more stable error throughout the rest of the window. This case is also notable because the most pronounced loss shaping occurred in the setting with longest predictive window, which is in line with our analysis of effect size by window length in Section~\ref{subsec:loss-reshaping}.

\subsection{Correlation Between Training and Test Errors.} We compute the Spearman correlation between mean stepwise errors computed on the training and test sets for models trained with constant constraint levels. Compared to the values reported for ERM in Table~\ref{tab:main-spearman}, we observe weaker correlations for our method. This can indicate that, while imposing a constraint on the training loss, our approach increasingly~\emph{overfits} to training data.

\begin{table}[!h]
\centering
\resizebox{.45\textwidth}{!}{
\begin{tabular}{ll|cccc}
\toprule
 &  & Autoformer & Informer & Reformer & Transformer \\
 &  &  &  &  &  \\
\midrule
\multirow[t]{4}{*}{\multirow{4}{*}{\rotatebox{90}{ECL}}} & 96 & 0.38 & 0.34 & 0.60 & 0.18 \\
 & 192 & 0.42 & 0.17 & 0.37 & -0.06 \\
 & 336 & 0.17 & 0.17 & -0.19 & -0.16 \\
 & 720 & 0.30 & 0.20 & 0.07 & 0.19 \\
\cline{1-6}
\multirow[t]{4}{*}{\multirow{4}{*}{\rotatebox{90}{Exchange}}} & 96 & 0.76 & 0.20 & 0.67 & -0.15 \\
 & 192 & 0.32 & 0.08 & 0.19 & -0.55 \\
 & 336 & 0.24 & -0.44 & 0.17 & 0.43 \\
 & 720 & 0.69 & -0.63 & 0.12 & -0.69 \\
\cline{1-6}
\multirow[t]{4}{*}{\multirow{4}{*}{\rotatebox{90}{Weather}}} & 96 & 0.01 & 0.23 & 0.14 & -0.30 \\
 & 192 & 0.43 & -0.08 & -0.43 & 0.33 \\
 & 336 & 0.46 & -0.22 & 0.55 & 0.25 \\
 & 720 & 0.83 & 0.45 & 0.69 & 0.64 \\
\cline{1-6}
\bottomrule
\end{tabular}
}
\caption{Spearman correlation for step-wise Mean Train and Test errors, for models trained with constant constraints.}\label{tab:main-spearman}
\end{table}

\subsection{Correlations Between Prediction Lengths} 
\begin{figure}
    \centering
    \includegraphics[width=0.5\textwidth]{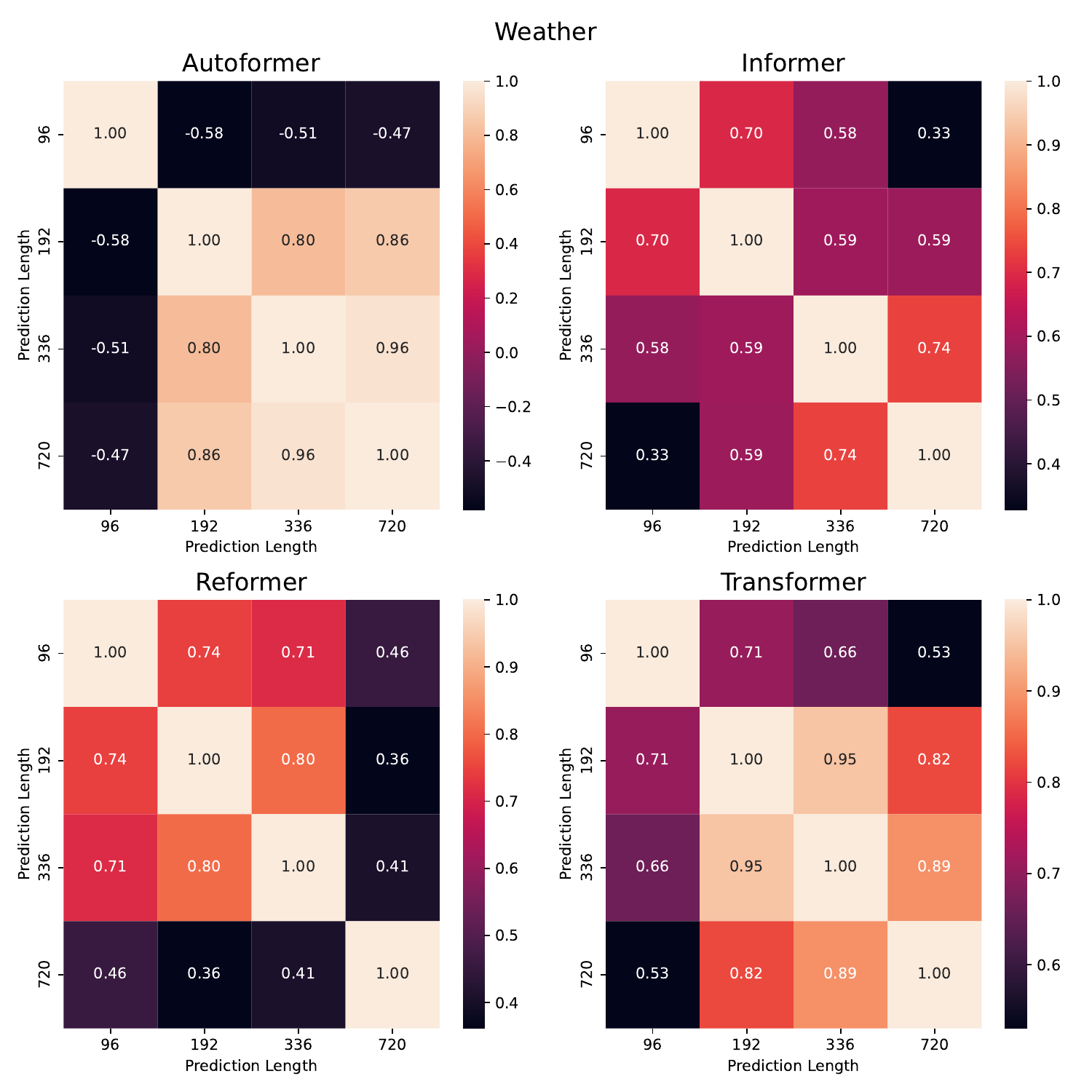}
    \centering
    \includegraphics[width=0.5\textwidth]{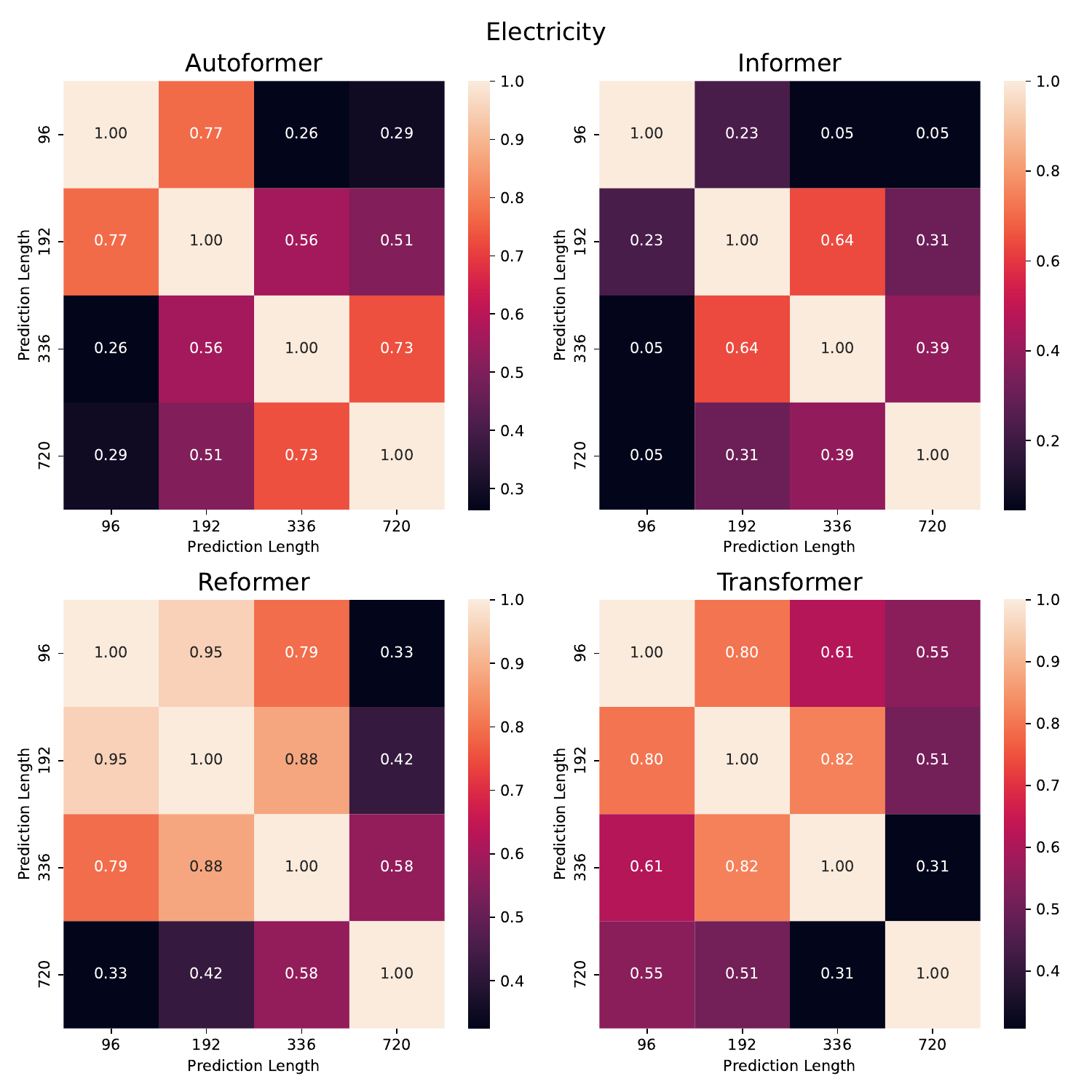}
    \caption{Pearson correlation between test step wise errors for weather and electricity datasets. Each heatmap shows pairwise correlations between all prediction lengths for a given model.}
    \label{fig:pred-len-corrs}
\end{figure}

In order to quantify the degree of similarity between error patterns across various prediction lengths, we compute Pearson correlation coefficients between step-wise errors, as shown in Figure~\ref{fig:pred-len-corrs}. We achieve this by first re-sampling shorter prediction windows using linear interpolation to match the number of steps. Notably, some setups present high correlations for several prediction lengths, which we interpret as the error distributions being similar.

\vspace{-0.12in}

\subsection{Ablation Analysis for Constant Constraints}\label{app-ablation}

\begin{table*}[hb!]
\centering
\begin{tabular}{lcccccc}
\toprule
\textbf{Model} & \textbf{Train 25\%} & \textbf{Train 50\%} & \textbf{Train 75\%} & \textbf{Val 25\%} & \textbf{Val 50\%} & \textbf{Val 75\%} \\
\midrule
Autoformer & 48.85 & 39.92 & 45.48 & 17.64 & 46.98 & 2.61 \\
FiLM & 29.88 & 29.64 & 29.61 & 29.44 & 29.18 & 28.87 \\
Informer & 0.85 & 1.54 & 1.51 & 56.48 & 60.50 & 60.50 \\
Nonstationary Transformer & -35.28 & -36.45 & -39.60 & -2.79 & 4.31 & 16.54 \\
Pyraformer & 7.82 & 14.63 & 14.63 & 15.89 & 21.22 & 23.40 \\
Reformer & 1.76 & 1.53 & 1.72 & -41.37 & -48.63 & -50.12 \\
Transformer & -0.73 & 0.24 & 0.20 & -2.46 & 53.18 & 54.38 \\
iTransformer & -11.75 & -13.02 & -13.91 & -15.52 & -8.76 & -8.74 \\
\bottomrule
\end{tabular}
\caption{MSE \% changes of constrained runs relative to ERM for varying values of $\epsilon$, in the Exchange Rate dataset, with window length 720.}
\label{tab:ablation_analysis}
\end{table*}

By examining the loss landscape across varying values of $\epsilon$ used on the Exchange Rate dataset, we find that our constrained method is not overly sensitive to the choice of hyperparameters. In  Table~\ref{tab:ablation_analysis}, we can observe each model's relative changes in MSE. Pyraformer, Reformer, Informer and Nonstationary Transformer exhibit a biphasic behavior: the loss shape changes drastically only after the $\epsilon$ is set to a parameter much higher than ERM's average training loss, which results in a trivial constraint. In the cases of FiLM and iTransformer, the shape remains relatively consistent across all values, while Autoformer exhibits larger variations. The Resilient approach is less sensitive to the constraint level, in some cases allowing for better generalization with the same value of epsilon compared to the constrained run.  

\begin{figure*}[ht!]
    \centering
    \includegraphics[width=0.90\textwidth]{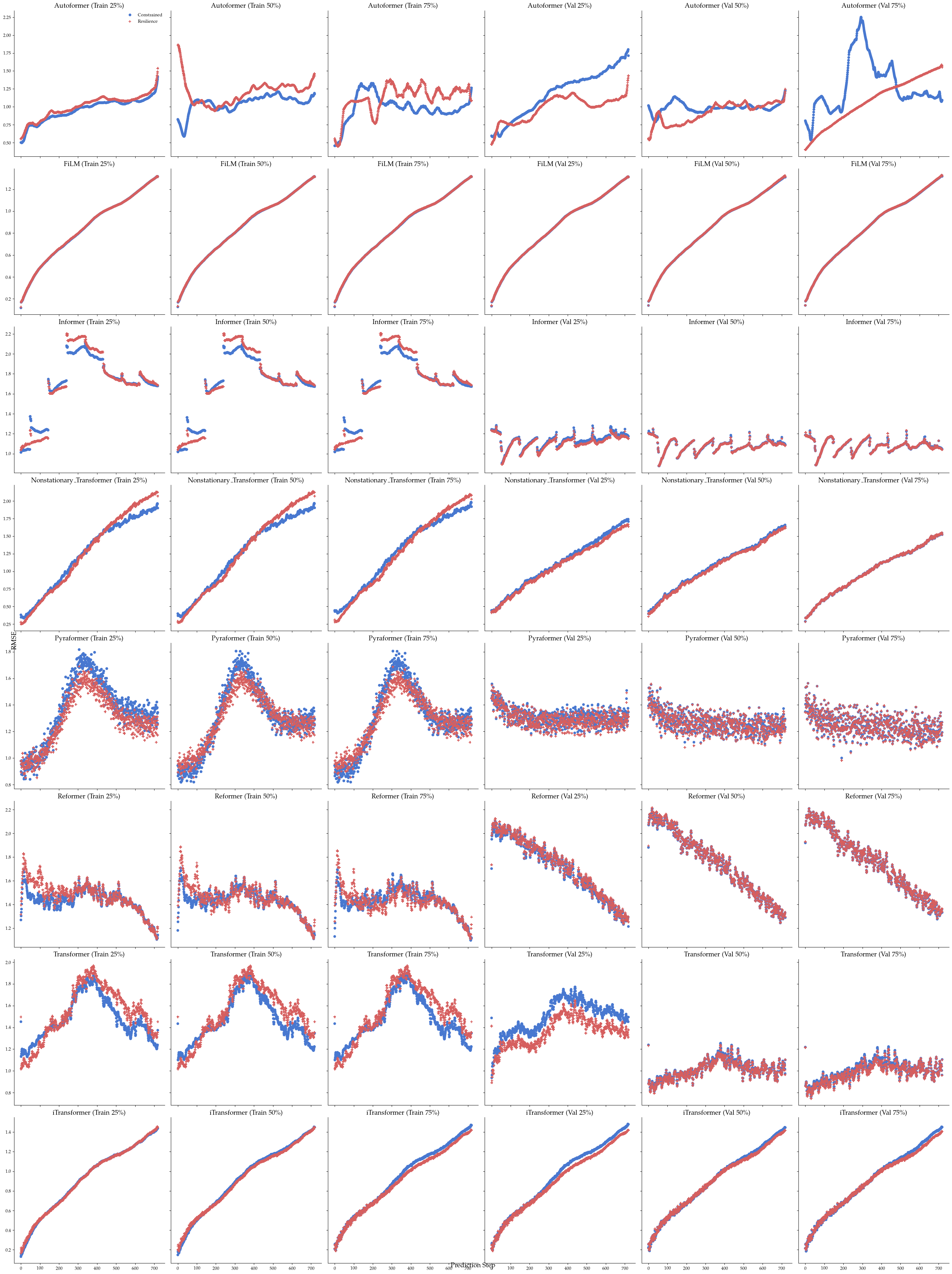}
    \centering
    
    \caption{Ablation plots for the $\epsilon$ parameter on the Exchange Rate dataset, with predictive window length of 720. Rows are different models, and columns are increasing values of $\epsilon$. Each plot shows test MSE of Constrained and Resilient runs.}
    \label{fig:ablation-epsilon}
\end{figure*}

Figure~\ref{fig:ablation-epsilon} illustrates the effect of the $\epsilon$ parameter on various models on the Exchange Rate dataset. We choose $\epsilon$ to be the quartiles of the train and test error taken from an ERM run. Notably, for several models, including the Informer, Reformer, and Transformer, the loss landscape exhibits relative invariance across similar $\epsilon$ values. This suggests a plateau in sensitivity within this parameter range. Conversely, in some cases, a critical $\epsilon$ threshold seems to exist, after which the shape significantly changes and then stabilizes again as we relax $\epsilon$ further.  Specifically, for Autoformer, Informer, and Transformer, optimal performance is achieved subsequent to a degree of $\epsilon$, underscoring the benefit of exploring multiple constraint levels.

We have shown that our method is reasonably robust to the choice of the constraint hyperparameter. However, we note that performing a parameter search with different values of $\epsilon$ may still be valuable. When the generalization gap is high, training with constraints higher than all of the training values is often more effective. For example, Reformer's test MSE is 2.9x higher than its train MSE, and Table~\ref{tab:ablation_analysis} shows its MSE can be reduced by up to 50.1\% when using validation-based threshold values, while training-based ones are not as effective.

\jp{Check if Reformer's generalization gap in exchange is higher than other models}
\ih{I think that the meaning of train based and val based is not clear here.}


\subsection{Outlier in STD histogram}\label{app-anomaly}

In Section~\ref{subsec:loss-reshaping}, we showed that on average, using constraints reduces Window STD. However, we identified some outliers in Figure~\ref{fig:histograms}. In one instance, the constrained Window STD is 371\% higher than ERM. It corresponds to Reformer on the Weather dataset with predictive window length of 720 steps, shown in Figure~\ref{fig:anomaly}. Closer inspection reveals that the reason for this increase is due to the constrained model finding a much lower MSE (constrained 0.788 versus ERM 0.963) in the first part of the window. Interestingly, the resilient model settles for a middle point, with a lower MSE than ERM but less variance increase than the constrained model.

\end{document}